# Integrating Learning from Examples into the Search for Diagnostic Policies

**Valentina Bayer-Zubek**                                    BAYER@CS.ORST.EDU
**Thomas G. Dietterich**                                     TGD@CS.ORST.EDU
*School of Electrical Engineering and Computer Science,*
*Dearborn Hall 102, Oregon State University,*
*Corvallis, OR 97331-3102 USA*

## Abstract

This paper studies the problem of learning diagnostic policies from training examples. A diagnostic policy is a complete description of the decision-making actions of a diagnostician (i.e., tests followed by a diagnostic decision) for all possible combinations of test results. An optimal diagnostic policy is one that minimizes the expected total cost, which is the sum of measurement costs and misdiagnosis costs. In most diagnostic settings, there is a tradeoff between these two kinds of costs.

This paper formalizes diagnostic decision making as a Markov Decision Process (MDP). The paper introduces a new family of systematic search algorithms based on the AO* algorithm to solve this MDP. To make AO* efficient, the paper describes an admissible heuristic that enables AO* to prune large parts of the search space. The paper also introduces several greedy algorithms including some improvements over previously-published methods. The paper then addresses the question of learning diagnostic policies from examples. When the probabilities of diseases and test results are computed from training data, there is a great danger of overfitting. To reduce overfitting, regularizers are integrated into the search algorithms. Finally, the paper compares the proposed methods on five benchmark diagnostic data sets. The studies show that in most cases the systematic search methods produce better diagnostic policies than the greedy methods. In addition, the studies show that for training sets of realistic size, the systematic search algorithms are practical on today's desktop computers.

## 1. Introduction

A patient arrives at a doctor's office complaining of symptoms such as fatigue, frequent urination, and frequent thirst. The doctor performs a sequence of measurements. Some of the measurements are simple questions (e.g., asking the patient's age, medical history, family history of medical conditions), others are simple tests (e.g., measure body mass index, blood pressure), and others are expensive tests (e.g., blood tests). After each measurement, the doctor analyzes what is known so far and decides whether there is enough information to make a diagnosis or whether more tests are needed. When making a diagnosis, the doctor must take into account the likelihood of each possible disease and the costs of misdiagnoses. For example, diagnosing a diabetic patient as healthy incurs the cost of aggravating the patient's medical condition and delaying treatment; diagnosing a healthy patient as having diabetes incurs the costs of unnecessary treatments. When the information that has been gathered is sufficiently conclusive, the doctor then makes a diagnosis.

We can formalize this diagnostic task as follows. Given a patient, the doctor can execute a set of $N$ possible measurements $x_1, \ldots, x_N$. When measurement $x_n$ is executed, the result is an observed value $v_n$. For example, if $x_1$ is "patient's age", then $v_1$ could be 36 (years). Each measurement $x_n$ has an associated cost $C(x_n)$. The doctor also can choose one of $K$ diagnosis actions. Diagnosis action $f_k$ diagnoses the patient as suffering from disease $k$. We will denote the correct diagnosis of the patient by $y$. The misdiagnosis cost of predicting disease $k$ when the correct diagnosis is $y$ is denoted by $MC(f_k, y)$.

The process of diagnosis consists of a sequence of decisions. In the starting state, no measurements or diagnoses have been made. We denote this by the empty set $\{\}$. Suppose that in this starting "knowledge state", the doctor chooses measurement $x_1$ and receives the result that $x_1 = 36$ at a cost of \$0.50. This is modeled as a transition to the knowledge state $\{x_1 = 36\}$ with a cost of $C(x_1) = 0.5$. Now suppose the doctor chooses $x_3$, which measures body mass index, and receives a result $x_3 = small$ at a cost of \$1. This changes the knowledge state to $\{x_1 = 36, x_3 = small\}$ at a cost of $C(x_3) = 1$. Finally, the doctor makes the diagnosis "healthy". Suppose that the correct diagnosis is $y = diabetes$. For illustrative purposes,[1] suppose that the cost of this misdiagnosis is $MC(healthy, diabetes) = \$100$. The diagnosis action terminates the process, with a total cost of $.5 + 1 + 100 = 101.5$.

We can summarize the decision-making process of the doctor in terms of a *diagnostic policy*, $\pi$. The diagnostic policy specifies for each possible knowledge state $s$, what action $\pi(s)$ to take, where the action can be one of the $N$ measurement actions or one of the $K$ diagnosis actions. Every diagnostic policy has an expected total cost, which depends on the joint probability distribution $P(x_1, \ldots, x_N, y)$ over the test results and the true disease of the patients and on the costs $C(x_n)$ and $MC(f_k, y)$. The optimal diagnostic policy minimizes this expected total cost by choosing the best tradeoff point between the cost of performing more measurements and the cost of misdiagnosis. Every measurement gathers information, which reduces the risk of a costly misdiagnosis. But every measurement incurs a measurement cost.

Diagnostic decision making is most challenging when the costs of measurement and misdiagnosis have similar magnitudes. If measurement costs are very small compared to misdiagnosis costs, then the optimal diagnostic policy is to measure everything and then make a diagnostic decision. Conversely, if misdiagnosis costs are very small compared to measurement costs, then the best policy is to measure nothing and just diagnose based on misdiagnosis costs and prior probabilities of the diseases.

Learning cost-sensitive diagnostic policies is important in many domains, from medicine to automotive troubleshooting to network fault detection and repair (Littman et al., 2004).

We note that this formulation of optimal diagnosis assumes that all costs can be expressed on a single numerical scale, that, although it need not correspond to economic cost, must support the principle of choosing actions by minimizing expected total cost. In medical diagnosis, there is a large body of work on methods for eliciting the patient's preferences and summarizing them as a utility or cost function (e.g., Lenert & Soetikno, 1997).

This paper studies the problem of learning diagnostic policies from training examples. We assume that we are given a representative set of complete training examples drawn from $P(x_1, \ldots, x_N, y)$ and that we are told the measurement costs and misdiagnosis costs. This kind of training data could be collected, for example, through a clinical trial in which all measurements were performed on all patients. Because of the costs involved in collecting such data, we assume that the training data sets will be relatively small (hundreds or a few thousands of patients; not millions). The goal of this paper is to develop learning algorithms for finding good diagnostic policies from such modest-sized training data sets. Unlike other work on test selection for diagnosis (Heckerman, Horvitz, & Middleton, 1993; van der Gaag & Wessels, 1993; Madigan & Almond, 1996; Dittmer & Jensen, 1997), we do not assume that a Bayesian network or influence diagram is provided; instead we directly learn a diagnostic policy from the data.

---

1. The true cost of misdiagnosing diabetes would depend on the age of the patient and the degree of progression of the disease, but in any case, it would probably be much higher than \$100.

This framework of diagnosis ignores several issues that we hope to address in future research. First, it assumes that each measurement action has no effect on the patient. Each measurement action is a pure observation action. In real medical and equipment diagnosis situations, some actions may also be attempted therapies or attempted repairs. These repairs may help cure the patient or fix the equipment, in addition to gathering information. Our approach does not handle attempted repair actions.

Second, this framework assumes that measurement actions are chosen and executed one-at-a-time and that the cost of an action does not depend on the order in which the actions are executed. This is not always true in medical diagnosis. For example, when ordering blood tests, the physician can choose to order several different tests as a group, which costs much less than if the tests are ordered individually.

Third, the framework assumes that the result of each measurement action is available before the diagnostician must choose the next action. In medicine, there is often a (stochastic) delay between the time a test is ordered and the time the results are available. Fragmentary results may arrive over time, which may lead the physician to order more tests before all previously-ordered results are available.

Fourth, the framework assumes that measurement actions are noise-free. That is, repeating a measurement action will obtain exactly the same result. Therefore once a measurement action is executed, it never needs to be repeated.

Fifth, the framework assumes that the results of the measurements have discrete values. We enforce this via a pre-processing discretization step.

These assumptions allow us to represent the doctor's knowledge state by the set of partial measurement results: $\{x_1 = v_1, x_3 = v_3, \dots\}$ and to represent the entire diagnostic process as a Markov Decision Process (MDP). Any optimal solution to this MDP provides an optimal diagnostic policy.

Given this formalization, there are conceptually two problems that must be addressed in order to learn good diagnostic policies. First, we must learn the joint probability distribution $P(x_1, \dots, x_N, y)$. Second, we must solve the resulting MDP for an optimal policy.

In this paper, we begin by addressing the second problem. We show how to apply the AO* algorithm to solve the MDP for an optimal policy. We define an admissible heuristic for AO* that allows it to prune large parts of the state space, so that this search becomes more efficient. This addresses the second conceptual problem.

However, instead of solving the first conceptual problem (learning the joint distribution $P(x_1, \dots, x_N, y)$) directly, we argue that the best approach is to integrate the learning process into the AO* search. There are three reasons to pursue this integration. First, by integrating learning into the search, we ensure that the probabilities computed during learning are the probabilities relevant to the task. If instead we had just separately learned some model of the joint distribution $P(x_1, \ldots, x_N, y)$, those probabilities would have been learned in a task-independent way, and long experience in machine learning has shown that it is better to exploit the task in guiding the learning process (e.g., Friedman & Goldszmidt, 1996; Friedman, Geiger, & Goldszmidt, 1997).

Second, by integrating learning into the search, we can introduce regularization methods that reduce the risk of overfitting. The more thoroughly a learning algorithm searches the space of possible policies, the greater the risk of overfitting the training data, which results in poor performance on new cases. The main contribution of this paper (in addition to showing how to model diagnosis as an MDP) is the development and careful experimental evaluation of several methods for regularizing the combined learning and AO* search process.

Third, the integration of learning with AO* provides additional opportunities to prune the AO* search and thereby improve the computational efficiency of the learning process. We introduce a pruning technique, called "statistical pruning", that simultaneously reduces the AO* search space and also regularizes the learning procedure.

In addition to applying the AO* algorithm to perform a systematic search of the space of diagnostic policies, we also consider greedy algorithms for constructing diagnostic policies. These algorithms are much more efficient than AO*, but we show experimentally that they give worse performance in several cases. Our experiments also show that AO* is feasible on all five diagnostic benchmark problems that we studied.

The remainder of the paper is organized as follows. First, we discuss the relationship between the problem of learning minimum cost diagnostic policies and previous work in cost-sensitive learning and diagnosis. In Section 3, we formulate this diagnostic learning problem as a Markov Decision Problem. Section 4 presents systematic and greedy search algorithms for finding good diagnostic policies. In Section 5, we take up the question of learning good diagnostic policies and describe our various regularization methods. Section 6 presents a series of experiments that measure the effectiveness and efficiency of the various methods on real-world data sets. Section 7 summarizes the contributions of the paper and discusses future research directions.

## 2. Relationship to Previous Research

The problem of learning diagnostic policies is related to several areas of previous research including cost-sensitive learning, test sequencing, and troubleshooting. We discuss each of these in turn.

### 2.1 Cost-Sensitive Learning

The term "cost-sensitive learning" denotes any learning algorithm that is sensitive to one or more costs. Turney (2000) provides an excellent overview. Cost-sensitive learning employs classification terminology in which a class is a possible outcome of the classification process. This corresponds in our case to the diagnosis. The forms of cost-sensitive learning most relevant to our work concern methods sensitive to misclassification costs, methods sensitive to measurement costs, and methods sensitive to both kinds of costs.

Learning algorithms sensitive to misclassification costs have received significant attention. In this setting, the learning algorithm is given (at no cost) the results of all possible measurements, $(v_1, \ldots, v_N)$. It must then make a prediction $\hat{y}$ of the class of the example, and it pays a cost $MC(\hat{y}, y)$ when the correct class is $y$. Important work in this setting includes the papers of Breiman et al. (1984), Pazzani et al. (1994), Fawcett and Provost (1997), Bradford et al. (1998), Domingos (Domingos, 1999), Zadrozny and Elkan (2001), and Provost and Fawcett (2001).

A few researchers in machine learning have studied application problems in which there is a cost for measuring each attribute (Norton, 1989; Nunez, 1991; Tan, 1993). In this setting, the goal is to minimize the number of misclassification errors while biasing the learning algorithm in favor of less-expensive attributes. From a formal point of view, this problem is ill-defined, because there is no explicit definition of an objective function that trades off the cost of measuring attributes against the number of misclassification errors. Nonetheless, several interesting heuristics were implemented and tested in these papers.

More recently, researchers have begun to consider both measurement and misclassification costs (Turney, 1995; Greiner, Grove, & Roth, 2002). The objective is identical to the one studied in this paper: to minimize the expected total cost of measurements and misclassifications. Both algorithms learn from data as well.

Turney developed ICET, an algorithm that employs genetic search to tune parameters that control a classification-tree learning algorithm. Each classification tree is built using a criterion that selects attributes greedily, based on their information gain and estimated costs. The measurement costs are adjusted in order to build different classification trees; these trees are evaluated on an internal holdout set using the real measurement and misclassification costs. The best set of measurement costs found by the genetic search is employed to build the final classification tree on the entire training data set.

Greiner et al.'s paper provides a PAC-learning analysis of the problem of learning an optimal diagnostic policy—provided that the policy makes no more than $L$ measurements, where $L$ is a fixed constant. Recall that $N$ is the total number of measurements. They prove that there exists an algorithm that runs in time polynomial in $N$, consumes a number of training examples polynomial in $N$, and finds a diagnostic policy that, with high probability, is close to optimal. Unfortunately, the running time and the required number of examples is exponential in $L$. In effect, their algorithm works by estimating, with high confidence, the transition probabilities and the class probabilities in states where at most $L$ of the values $x_1 = v_1, \ldots, x_N = v_N$ have been observed. Then the value iteration dynamic programming algorithm is applied to compute the best diagnostic policy with at most $L$ measurements. In theory, this works well, but it is difficult to convert this algorithm to work in practice. This is because the theoretical algorithm chooses the space of possible policies and then computes the number of training examples needed to guarantee good performance, whereas in a real setting, the number of available training examples is fixed, and it is the space of possible policies that must be adapted to avoid overfitting.

## 2.2 Test Sequencing

The field of electronic systems testing has formalized and studied a problem called the *test sequencing problem* (Pattipati & Alexandridis, 1990). An electronic system is viewed as being in one of $K$ possible states. These states include one fault-free state and $K - 1$ faulty states. The relationship between tests (measurements) and system states is specified as a binary diagnostic matrix which tells whether test $x_n$ detects fault $f_i$ or not. The probabilities of the different system states $y$ are specified by a known distribution $P(y)$.

A test sequencing policy performs a series of measurements to identify the state of the system. In test sequencing, it is assumed that the measurements are sufficient to determine the system state with probability 1. The objective is to find the test sequencing policy that achieves this while minimizing the expected number of tests. Hence, misdiagnosis costs are irrelevant, because the test sequencing policy must guarantee zero misdiagnoses. Several heuristics for AO* have been applied to compute the optimal test sequencing policy (Pattipati & Alexandridis, 1990).

The test sequencing problem does not involve learning from examples. The required probabilities are provided by the diagnostic matrix and the fault distribution $P(y)$.

### 2.3 Troubleshooting

Another task related to our work is the task of troubleshooting (Heckerman, Breese, & Rommelse, 1994). Troubleshooting begins with a system that is known to be functioning incorrectly and ends when the system has been restored to a correctly-functioning state. The troubleshooter has two kinds of actions: pure observation actions (identical to our measurement actions) and repair actions (e.g., removing and replacing a component, replacing batteries, filling the gas tank, rebooting the computer, etc.). Each action has a cost, and the goal is to find a troubleshooting policy that minimizes the expected cost of restoring the system to a correctly-functioning state.

Heckerman et al. (1994, 1995) show that for the case where the only actions are pure repair actions and there is only one broken component, there is a very efficient greedy algorithm that computes the optimal troubleshooting policy. They incorporate pure observation actions via a one-step value of information (VOI) heuristic. According to this heuristic, they compare the expected cost of a repair-only policy with the expected cost of a policy that makes exactly one observation action and then executes a repair-only policy. If an observe-once-and-then-repair-only policy is better, they execute the chosen observation action, obtain the result, and then again compare the best repair-only policy with the best observe-once-and-then-repair-only policy. Below, we define a variant of this VOI heuristic and compare it to the other greedy and systematic search algorithms developed in this paper.

Heckerman et al. consider only the case where the joint distribution $P(x_1, \ldots, x_N, y)$ is provided by a known Bayesian network. To convert their approach into a learning approach, they could first learn the Bayesian network and then compute the troubleshooting policy. But we suspect that an approach that integrates the learning of probabilities into the search for good policies—along the lines described in this paper—would give better results. Exploring this question is an important direction for future research.

## 3. Formalizing Diagnosis as a Markov Decision Problem

The process of diagnosis is a sequential decision making process. After every decision, the diagnostician must decide what to do next (perform another measurement, or terminate by making a diagnosis). This can be modeled as a Markov Decision Problem (MDP).

An MDP is a mathematical model for describing the interaction of an agent with an environment. An MDP is defined by a set of states $S$ (including the start state), an action set $A$, the transition probabilities $P_{tr}(s'|s, a)$ of moving from state $s$ to state $s'$ after executing action $a$, and the (expected immediate) costs $C(s, a, s')$ associated with these transitions. Because the state representation contains all the relevant information for future decisions, it is said to exhibit the *Markov property*.

A policy $\pi$ maps states into actions. The value of a state $s$ under policy $\pi$, $V^\pi(s)$, is the expected sum of future costs incurred when starting in state $s$ and following $\pi$ afterwards (Sutton & Barto, 1999, chapter 3). The value function $V^\pi$ of a policy $\pi$ satisfies the following recursive relationship, known as the *Bellman equation for $V^\pi$*:

$$V^\pi(s) = \sum_{s' \in S} P_{tr}(s'|s, \pi(s)) \times [C(s, \pi(s), s') + V^\pi(s')], \forall \pi, \forall s. \qquad (1)$$

This can be viewed as a *one-step lookahead* from state $s$ to each of the next states $s'$ reached after executing $\pi(s)$. Given a policy $\pi$, the value of state $s$ can be computed from the value of its successor states $s'$, by adding the expected costs of the transitions, then weighting them by the transition probabilities.

Solving the MDP means finding a policy with the smallest value. Such a policy is called the *optimal policy $\pi^*$*, and its value is the *optimal value function $V^*$*. *Value iteration* is an algorithm that solves MDPs by iteratively computing $V^*$ (Puterman, 1994).

The problem of learning diagnostic policies can be represented as an MDP. We first define the actions of this MDP, then the states, and finally the transition probabilities and costs. *All costs are positive.*

As discussed above, we assume that there are $N$ measurement actions (tests) and $K$ diagnosis actions. Measurement action $n$ (denoted $x_n$) returns the value of attribute $x_n$, which we assume is a discrete variable with $V_n$ possible values. Diagnosis action $k$ (denoted $f_k$) is the act of predicting that the correct diagnosis of the example is $k$. An action (measurement or diagnosis) is denoted by $a$.

In our diagnostic setting, a case is completely described by the results of all $N$ measurement actions and the correct diagnosis $y$: $(v_1, \ldots, v_N, y)$. In our framework, each case is drawn independently according to an (unknown) joint distribution $P(x_1, \ldots, x_N, y)$. Once a case is drawn, all the values defining it stay constant. Test $x_n$ reveals to the diagnostic agent the value $x_n = v_n$ of this case. As a consequence, once a case has been drawn, the order in which the tests are performed does not change the values that will be observed. That is, the joint distribution $P(x_i = v_i, x_j = v_j)$ is independent of the order of the tests $x_i$ and $x_j$.

It follows that we can define the state of the MDP as the set of all attribute-value pairs observed thus far. This state representation has the Markov property because it contains all relevant past information. There is a unique *start state*, $s_0 = \{\}$, in which no attributes have been measured. The set of all states $S$ contains one state for each possible combination of measured attributes, as found in the training data. Each training example provides evidence for the reachability of $2^N$ states. The set $A(s)$ of actions executable in state $s$ consists of those attributes not yet measured plus all of the diagnosis actions.

We also define a special *terminal state $s_f$*. Every diagnosis action makes a transition to $s_f$ with probability 1 (i.e., once a diagnosis is made, the task terminates). By definition, no actions are executable in the terminal state, and its value function is zero. Note that the terminal state is always reached, because there are only finitely-many measurement actions after which a diagnosis action must be executed.

We now define the transition probabilities and the immediate costs of the MDP. For measurement action $x_n$ executed in state $s$, the result state will be $s' = s \cup \{x_n = v_n\}$, where $v_n$ is the observed value of $x_n$. The expected cost of this transition is denoted $C(x_n)$, since we assume it depends only on which measurement action $x_n$ is executed, and not on the state in which it is executed nor the resulting value $v_n$ that is observed. The probability of this transition is $P_{tr}(s'|s, x_n) = P(x_n = v_n|s)$.

The misdiagnosis cost of diagnosis action $f_k$ depends on the correct diagnosis $y$ of the example. Let $MC(f_k, y)$ be the misdiagnosis cost of guessing diagnosis $k$ when the correct diagnosis is $y$. Because the correct diagnosis $y$ of an example is not part of the state representation, the cost of a diagnosis action (which depends on $y$) performed in state $s$ must be viewed as a random variable whose value is $MC(f_k, y)$ with probability $P(y|s)$, which is the probability that the correct diagnosis is $y$ given the current state $s$. Hence, our MDP has a stochastic cost function for diagnosis actions. This does not lead to any difficulties, because all that is required to compute the optimal policy for an MDP is the *expected* cost of each action. In our case, the expected cost of diagnosis action $f_k$ in state $s$ is

$$C(s, f_k) = \sum_y P(y|s) \cdot MC(f_k, y), \tag{2}$$

which is independent of $y$.

For uniformity of notation, we will write the expected immediate cost of action $a$ in state $s$ as $C(s, a)$, where $a$ can be either a measurement action or a diagnosis action.

For a given start state $s_0$, the diagnostic policy $\pi$ is a *decision tree* (Raiffa, 1968). Figure 1 illustrates a simple example of a diagnostic policy. The root is the starting state $s_0 = \{\}$. Each node is labeled with a state $s$ and a corresponding action $\pi(s)$. If the action is a measurement action, $x_n$, the possible results are the different possible observed values $v_n$, leading to children nodes. If the action is a diagnosis action, $f_k$, the possible results are the diagnoses $y$. If $\pi(s)$ is a measurement action, the node is called an *internal node*, and if $\pi(s)$ is a diagnosis action, the node is called a *leaf node*. Each branch in the tree is labeled with its probability of being followed (conditioned on reaching its parent node). Each node $s$ is labeled with $V^\pi(s)$, the expected total cost of executing the diagnostic policy starting at node $s$. Notice that the value of a leaf is the expected cost of diagnosis, $C(s, f_k)$.

The fact that a diagnostic policy is a decision tree is potentially confusing, because a similar data structure, the classification tree (often also called a decision tree), has been the focus of so much work in the machine learning literature (e.g., Quinlan, 1993). It is important to remember that whereas the evaluation criterion for a classification tree is the misclassification error rate, the evaluation criterion for a decision tree *diagnostic policy* is the expected total cost of diagnosis. One way of clarifying this difference is to note that a given classification tree can be transformed into many equivalent classification trees by changing the order in which the tests are performed (see Utgoff's work on tree manipulation operators, Utgoff, 1989). These equivalent classifiers all implement the same classification function $y = f(x_1, \ldots, x_N)$. But if we consider these "equivalent" trees as diagnostic policies, they will have different expected total diagnosis costs, because tests closer to the root of the tree will be executed more often, so their measurement costs will make a larger contribution to the total diagnosis cost. For example, the policy $\pi$ in Figure 1 first performs a cheap test, BMI. This policy has a value of 28.99. The tree $\pi_2$ in Figure 2 makes the same classification decisions (with an error rate of 19%), but it first tests Insulin, which is more expensive, and this increases the policy value to 40.138.

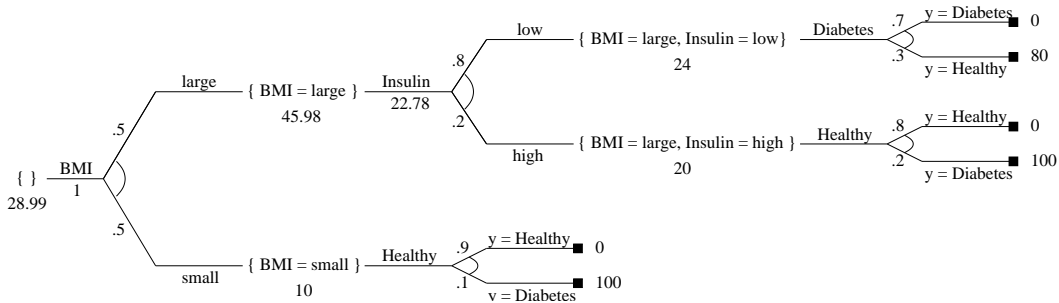

Figure 1: An example of diagnostic policy $\pi$ for diabetes. Body Mass Index (BMI) is tested first. If it is small, a Healthy diagnosis is made. If BMI is large, Insulin is tested before making a diagnosis. The costs of measurements (BMI and Insulin) are written below the name of the variable. The costs of misdiagnoses are written next to the solid squares. Probabilities are written on the branches. The values of the states are written below each state. The value of the start state, $V^\pi(s_0) = 28.99$, can be computed in a single sweep, starting at the leaves, as follows. First the expected costs of the diagnosis actions are computed (e.g., the upper-most Diabetes diagnosis action has an expected cost of $0.7 \times 0 + 0.3 \times 80 = 24$). Then the value of the Insulin subtree is computed as the cost of measuring Insulin $(22.78) + 0.8 \times 24 + 0.2 \times 20 = 45.98$. Finally, the value of the whole tree is computed as the cost of measuring BMI $(1) + 0.5 \times 45.98 + 0.5 \times 10 = 28.99$.

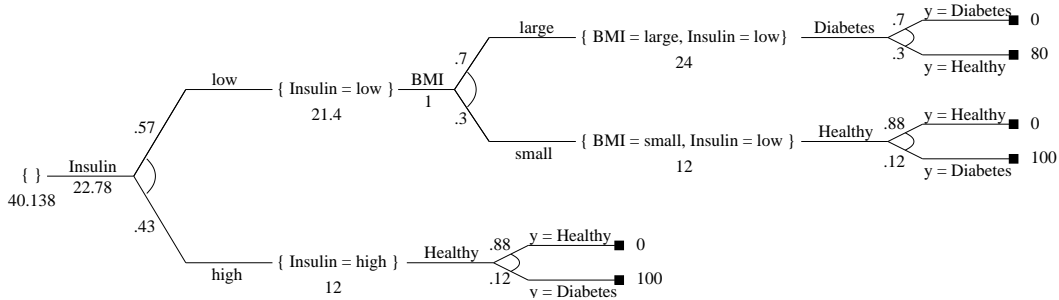

Figure 2: Another diagnostic policy $\pi_2$, making the same classification decisions as $\pi$ in Figure 1, but with a changed order of tests, and therefore with a different policy value.

## 4. Searching for Good Diagnostic Policies

We now consider systematic and greedy search algorithms for computing diagnostic policies. In this section, we will assume that all necessary probabilities are known. We defer the question of learning those probabilities to Section 5. We note that this is exactly what all previous uses of AO* have done. They have always assumed that the required probabilities and costs were known.

Given the MDP formulation of the diagnostic process, we could proceed by constructing the entire state space and then applying dynamic programming algorithms (e.g., value iteration or policy iteration) to find the optimal policy. However, the size of the state space is exponential: given $N$ measurement actions, each with $V$ possible outcomes, there are $(V + 1)^N + 1$ states in the MDP (counting the special terminal state $s_f$, and taking into account that each measurement may not have been performed yet). We seek search algorithms that only consider a small fraction of this huge space. In this section, we will study two general approaches to dealing with this combinatorial explosion of states: systematic search using the AO* algorithm and various greedy search algorithms.

### 4.1 Systematic Search

When an MDP has a unique start state and no (directed) cycles, the space of policies can be represented as an AND/OR graph (Qi, 1994; Washington, 1997; Hansen, 1998). An AND/OR graph is a directed acyclic graph that alternates between two kinds of nodes: AND nodes and OR nodes. Each OR node represents a state $s$ in the MDP state space. Each child of an OR node is an AND node that represents one possible action $a$ executed in state $s$. Each child of an AND node is an OR node that represents a state $s'$ that results from executing action $a$ in state $s$. Figure 3 shows an example of an AND/OR graph for a diabetes diagnosis problem with three tests (BMI, Glucose, and Insulin) and two diagnosis actions (Diabetes and Healthy).

In our diagnostic setting, the root OR node corresponds to the starting state $s_0 = \{\}$. Each OR node $s$ has one AND child $(s, x_n)$ for each measurement action (test) $x_n$ that can be executed in $s$. Each OR node could also have one child for each possible diagnosis action $f_k$ that could be performed in $s$, but to save time and memory, we include only the one diagnosis action $f_k$ that has the minimum expected cost. We will denote this by $f_{best}$. Each time an OR node is created, an AND child for $f_{best}$ is created immediately. This leaf AND node stores the action-value function $Q(s, f_{best}) = C(s, f_{best})$. Note that multiple paths from the root may lead to the same OR node, by changing the order of the tests.

In our implementation, each OR node stores a representation of the state $s$, a current policy $\pi(s)$ which specifies a test or a diagnosis action, and a current value function $V^\pi(s)$. Each AND node $(s, x_n)$ stores a probability distribution over the outcomes of $x_n$, and an action-value function $Q^\pi(s, x_n)$, the expected cost of measuring $x_n$ and then continuing with policy $\pi$.

Every possible policy $\pi$ corresponds to a subtree of the full AND/OR graph. Each OR node $s$ in this subtree (starting at the root) contains only the one AND child $(s, a)$ corresponding to the action $a = \pi(s)$ chosen by policy $\pi$.

The AO* algorithm (Nilsson, 1980) computes the optimal policy for an AND/OR graph. AO* is guided by a heuristic function. We describe the heuristic function in terms of state-action pairs, $h(s, a)$, instead of in terms of states. The heuristic function is *admissible* if $h(s, a) \leq Q^*(s, a)$ for all states $s$ and actions $a$. This means that $h$ underestimates the total cost of executing action $a$ in state $s$ and following the optimal policy afterwards. The admissible heuristic allows the AO* algorithm to safely ignore an action $a'$ if there is another action $a$ for which it is known that $Q^*(s, a) < h(s, a')$. Under these conditions, $(s, a')$ cannot be part of any optimal policy.

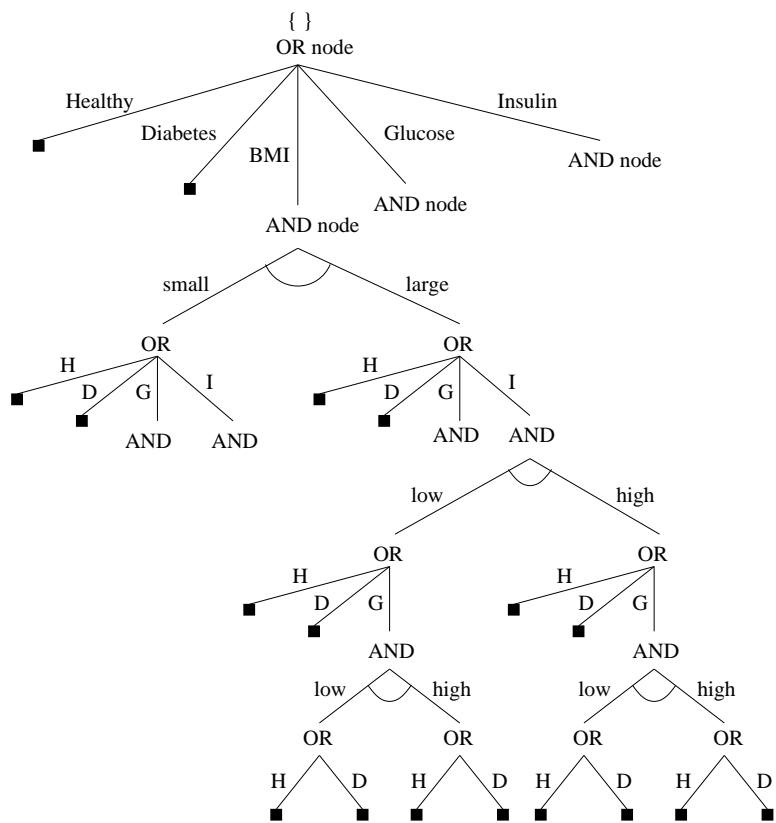

Figure 3: An example of an AND/OR graph. The root OR node corresponds to the state $s_0 = \{\}$. There is a child AND node for each of the test actions (BMI, Glucose and Insulin), and also for the diagnosis actions (Healthy and Diabetes). The choice of the BMI test in the root node leads to the AND node $(s_0, BMI)$, which specifies the expectation over the outcomes of the test BMI (small and large). If BMI is small, the child of AND node $(s_0, BMI)$ is the OR node with state $\{BMI = small\}$; in this OR node, there is a choice among the actions Healthy, Diabetes, Glucose and Insulin.

The AO* search begins with an AND/OR graph containing only the root node. It then repeats the following steps: In the current best policy, it selects an AND node to expand; it expands it (*expanding* an AND node creates its children OR nodes); and then it recomputes (bottom-up) the optimal value function and policy of the revised graph. The algorithm terminates when the best policy has no unexpanded AND nodes (in other words, the leaf

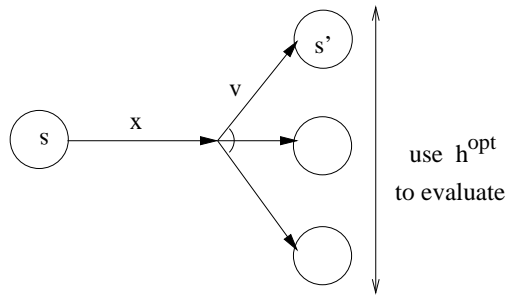

Figure 4: $Q^{opt}(s, x)$ for unexpanded AND node $(s, x)$ is computed using one-step lookahead and $h^{opt}$ to evaluate the resulting states $s'$. $x$ is an attribute not yet measured in state $s$, and $v$ is one of its values.

OR nodes of the policy specify diagnosis actions, so this policy is a *complete* diagnostic policy).

During AO* search, we maintain two policies, whose actions and value functions are stored in the nodes of the AND/OR graph. We call the first policy the *optimistic policy*, $\pi^{opt}$. As we show below, its value function $V^{opt}$ is a lower bound on the optimal value function $V^*$. This is the policy that appears in Nilsson's original description of AO*, and it provides enough information to compute an optimal policy $\pi^*$ (Martelli & Montanari, 1973). During the search, the optimistic policy $\pi^{opt}$ is an incomplete policy, because it includes some unexpanded AND nodes; when $\pi^{opt}$ becomes a complete policy, it is in fact an optimal policy.

The second policy that we maintain is called the *realistic policy*, $\pi^{real}$. We will show that its value function, $V^{real}$, is an upper bound on the optimal value function $V^*$. The realistic policy is always a complete policy, so it is executable after each iteration of AO*. By maintaining the realistic policy, AO* becomes an anytime algorithm.

We now define these two policies in more detail and introduce our admissible heuristic.

### 4.1.1 Admissible Heuristic

Our admissible heuristic provides an optimistic estimate, $Q^{opt}(s, x)$, of the expected cost of an unexpanded AND node $(s, x)$. It is based on an incomplete two-step lookahead search (see Figure 4). The first step of the lookahead search computes $Q^{opt}(s, x) = C(s, x) + \sum_{s'} P_{tr}(s'|s, x) \cdot h^{opt}(s')$. Here $s'$ iterates over the states resulting from measuring test $x$. The second step of the lookahead is defined by the function $h^{opt}(s') = \min_{a' \in A(s')} C(s', a')$, which is the minimum over the cost of the diagnosis action $f_{best}$ and the cost of each of the remaining tests $x'$ in $s'$. That is, rather than considering the states $s''$ that would result from measuring $x'$, we only consider the cost of measuring $x'$ itself. It follows immediately that $h^{opt}(s') \le V^*(s'), \forall s'$, because $C(s', x') \le Q^*(s', x') = C(s', x') + \sum_{s''} P_{tr}(s''|s', x') \cdot V^*(s'')$. The key thing to notice is that the cost of a single measurement $x'$ is less than or equal to the cost of any policy that begins by measuring $x'$, because the policy must pay the cost of at least one more action (diagnosis or measurement) before entering the terminal state $s_f$. Consequently, $Q^{opt}(s, x) \le Q^*(s, x)$, so $Q^{opt}$ is an admissible heuristic for state $s$ and action $x$.

#### 4.1.2 OPTIMISTIC VALUES AND OPTIMISTIC POLICY

The definition of the optimistic action-value value $Q^{opt}$ can be extended to all AND nodes in the AND/OR graph through the following recursion:

$$Q^{opt}(s,a) = \begin{cases} C(s,a), \text{ if } a = f_k \text{ (a diagnosis action)} \\ C(s,a) + \sum_{s'} P_{tr}(s'|s,a) \cdot h^{opt}(s'), \text{ if } (s,a) \text{ is unexpanded} \\ C(s,a) + \sum_{s'} P_{tr}(s'|s,a) \cdot V^{opt}(s'), \text{ if } (s,a) \text{ is expanded,} \end{cases} \quad (3)$$

where $V^{opt}(s) \overset{\text{def}}{=} \min_{a \in A(s)} Q^{opt}(s,a)$. Recall that $A(s)$ consists of all attributes not yet measured in $s$ and all diagnosis actions.

The optimistic policy is $\pi^{opt}(s) = \operatorname{argmin}_{a \in A(s)} Q^{opt}(s,a)$. Every OR node $s$ stores its optimistic value $V^{opt}(s)$ and policy $\pi^{opt}(s)$, and every AND node $(s,a)$ stores $Q^{opt}(s,a)$. Theorem 4.1 proves that $Q^{opt}$ and $V^{opt}$ form an admissible heuristic. The proofs for all theorems in this paper appear in the thesis of Bayer-Zubek (2003).

**Theorem 4.1** *For all states $s$ and all actions $a \in A(s)$, $Q^{opt}(s,a) \leq Q^*(s,a)$, and $V^{opt}(s) \leq V^*(s)$.*

#### 4.1.3 REALISTIC VALUES AND REALISTIC POLICY

In the current graph constructed by AO*, suppose that we delete all unexpanded AND nodes $(s,a)$. We call the resulting graph the *realistic* graph, because every leaf node will select a diagnosis action. The optimal policy computed from this graph is called the *realistic policy*, $\pi^{real}$. It is a complete policy leaves specify diagnosis actions of minimum expected misdiagnosis cost.

Every OR node $s$ stores the realistic value $V^{real}(s)$ and policy $\pi^{real}(s)$, and every AND node $(s,a)$ stores a realistic action-value value, $Q^{real}(s,a)$. For $a \in A(s)$, define

$$Q^{real}(s,a) = \begin{cases} C(s,a), \text{ if } a = f_k \text{ (a diagnosis action)} \\ C(s,a) + \sum_{s'} P_{tr}(s'|s,a) \cdot V^{real}(s'), \text{ if } (s,a) \text{ is expanded} \\ \text{ignore, if } (s,a) \text{ is unexpanded} \end{cases} \quad (4)$$

and $V^{real}(s) = \min_{a \in A'(s)} Q^{real}(s,a)$, where the set $A'(s)$ is $A(s)$ without the unexpanded actions. The realistic policy is $\pi^{real}(s) = \operatorname{argmin}_{a \in A'(s)} Q^{real}(s,a)$.

**Theorem 4.2** *The realistic value function $V^{real}$ is an upper bound on the optimal value function: $V^*(s) \leq V^{real}(s), \forall s$.*

#### 4.1.4 SELECTING A NODE FOR EXPANSION

In the current optimistic policy $\pi^{opt}$, we choose to expand the unexpanded AND node $(s, \pi^{opt}(s))$ with the largest impact on the root node. This is defined as

$$\operatorname{argmax}_s \left[ V^{real}(s) - V^{opt}(s) \right] \cdot P_{reach}(s|\pi^{opt}),$$

where $P_{reach}(s|\pi^{opt})$ is the probability of reaching state $s$ from the start state while following policy $\pi^{opt}$. The difference $V^{real}(s) - V^{opt}(s)$ is an upper bound on how much the value of state $s$ could change if $\pi^{opt}(s)$ is expanded.

The rationale for this selection is based on the observation that AO* terminates when $V^{opt}(s_0) = V^{real}(s_0)$. Therefore, we want to expand the node that makes the biggest step toward this goal.

### 4.1.5 OUR IMPLEMENTATION OF AO* (HIGH LEVEL)

Our implementation of AO* is the following:

> **repeat**
>     select an AND node $(s, a)$ to expand (using $\pi^{opt}, V^{opt}, V^{real}$).
>     expand $(s, a)$.
>     do bottom-up updates of $Q^{opt}, V^{opt}, \pi^{opt}$ and of $Q^{real}, V^{real}, \pi^{real}$.
> **until** there are no unexpanded nodes reachable by $\pi^{opt}$.

The updates of value functions are based on one-step lookaheads (Equations 3 and 4), using the value functions of the children. At each iteration, we start from the newly expanded AND node $(s, a)$, compute its $Q^{opt}(s, a)$ and $Q^{real}(s, a)$, then compute $V^{opt}(s), \pi^{opt}(s)$, $V^{real}(s)$, and $\pi^{real}(s)$ in its parent OR node, and propagate these changes up in the AND/OR graph all the way to the root. Full details on our implementation of AO* appear in the thesis of Bayer-Zubek (2003).

As more nodes are expanded, the optimistic values $V^{opt}$ increase, becoming tighter lower bounds to the optimal values $V^*$, and the realistic values $V^{real}$ decrease, becoming tighter upper bounds. $V^{opt}$ and $V^{real}$ converge to the value of the optimal policy: $V^{opt}(s) = V^{real}(s) = V^*(s)$, for all states $s$ reached by $\pi^*$.

The admissible heuristic avoids exploring expensive parts of the AND/OR graph; indeed, when $V^{real}(s) < Q^{opt}(s, a)$, action $a$ does not need to be expanded (this is a *heuristic cutoff*). Initially, $V^{real}(s) = C(s, f_{best})$, and this explains why measurement costs that are large relative to misdiagnosis costs produce many cutoffs.

## 4.2 Greedy Search

Now that we have considered the AO* algorithm for systematic search, we turn our attention to several greedy search algorithms for finding good diagnostic policies. Greedy search algorithms grow a decision tree starting at the root, with state $s_0 = \{\}$. Each node in the tree corresponds to a state $s$ in the MDP, and it stores the corresponding action $a = \pi(s)$ chosen by the greedy algorithm. The children of node $s$ correspond to the states that result from executing action $a$ in state $s$. If a diagnosis action $f_k$ is chosen in state $s$, then the node has no children in the decision tree (it is a leaf node).

All of the greedy algorithms considered in this paper share the same general template, which is shown as pseudo-code in Table 1. At each state $s$, the greedy algorithm performs a limited lookahead search and then commits to the choice of an action $a$ to execute in $s$, which thereby defines $\pi(s) = a$. It then generates child nodes corresponding to the states that could result from executing action $a$ in state $s$. The algorithm is then invoked recursively on each of these child nodes.

Once a greedy algorithm has committed to $x_n = \pi(s)$, that choice is final. Note however, that some of our regularization methods may prune the policy by replacing a measurement action (and its descendents) with a diagnosis action. In general, greedy policies are not optimal, because they do not perform a complete analysis of the expected total cost of

Table 1: The Greedy search algorithm. Initially, the function Greedy() is called with the start state $s_0$.

**function** Greedy(state $s$) **returns** a policy $\pi$ (in the form of a decision tree).

(1) **if** (*stopping conditions* are not met)
(2)     *select measurement action $x_n$ to execute*
        set $\pi(s) := x_n$
        for each resulting value $v_n$ of the test $x_n$ add the subtree
            Greedy(state $s \cup \{x_n = v_n\}$)
    **else**
(3)     *select diagnosis action $f_{best}$*, set $\pi(s) := f_{best}$.

executing $x_n$ in $s$ before committing to an action. Nevertheless, they are efficient because of their greediness.

In the following discussion, we describe several different greedy algorithms. We define each one by describing how it refines the numbered lines in the template of Table 1.

### 4.2.1 InfoGainCost Methods

InfoGainCost methods are inspired by the C4.5 algorithm for constructing classification trees (Quinlan, 1993). C4.5 chooses the attribute $x_n$ with the highest conditional mutual information with the class labels in the training examples. In our diagnostic setting, the analogous criterion is to choose the measurement action that is most predictive of the correct diagnosis. Specifically, let $x_n$ be a proposed measurement action, and define $P(x_n, y|s)$ to be the joint distribution of $x_n$ and the correct diagnosis $y$ conditioned on the information that has already been collected in state $s$. The conditional mutual information between $x_n$ and $y$, $I(x_n; y|s)$, is defined as

$$
\begin{aligned}
I(x_n; y|s) &= H(y|s) - H(y|s, x_n) \\
&= H(y|s) - \sum_{v_n} P(x_n = v_n|s) \cdot H(y|s \cup \{x_n = v_n\})
\end{aligned}
$$

where $H(y) = \sum_y -P(y) \log P(y)$ is the Shannon entropy of random variable $y$.

The mutual information is also called the information gain, because it quantifies the average amount of information we gain about $y$ by measuring $x_n$.

The InfoGainCost methods penalize the information gain by dividing it by the cost of the test. Specifically, they choose the action $x_n$ that maximizes $I(x_n; y|s)/C(x_n)$. This criterion was introduced by Norton (1989). Other researchers have considered various monotonic transformations of the information gain prior to dividing by the measurement cost (Tan, 1993; Nunez, 1991). This defines step (2) of the algorithm template.

All of the InfoGainCost methods employ the stopping conditions defined in C4.5. The first stopping condition applies if $P(y|s)$ is 1 for some value $y = k$. In this case, the diagnosis action is chosen to be $f_{best} = k$. The second stopping condition applies if no more

measurement actions are available (i.e., all tests have been performed). In this case, the diagnosis action is set to the most likely diagnosis: $f_{best} := \text{argmax}_y P(y|s)$.

Notice that the InfoGainCost methods do not make any use of the misdiagnosis costs $MC(f_k, y)$.

### 4.2.2 MODIFIED INFOGAINCOST METHODS (MC+INFOGAINCOST)

We propose extending the InfoGainCost methods so that they consider misdiagnosis costs in the stopping conditions. Specifically, in step (3), the MC+InfoGainCost methods set $f_{best}$ to be the diagnosis action with minimum expected cost:

$$\pi(s) := f_{best} = \underset{f_k}{\text{argmin}} \sum_y P(y|s) \cdot MC(f_k, y).$$

### 4.2.3 ONE-STEP VALUE OF INFORMATION (VOI)

While the previous greedy methods either ignore the misdiagnosis costs or only consider them when choosing the final diagnosis actions, the VOI approach considers misdiagnosis costs (and measurement costs) at each step.

Traditionally, the value of information of a measurement is defined as the difference between the expected value of the best action after performing the measurement and the expected value of the best action before performing the measurement. Since our objective is cost minimization, we need to reverse the sign in the above definition. However, we still keep the notation VOI instead of cost of information. Instead of taking into account all future decisions, we make a greedy approximation to VOI, called one-step VOI, in which we only consider the cost of the best diagnosis action before and after performing the measurement $x_n$ in state $s$:

$$\begin{aligned}
\text{1-step-VOI}(s, x_n) \quad = \quad & \min_{f_k} \sum_y P(y|s) \cdot MC(f_k, y) \\
& - \sum_{v_n} P(x_n = v_n | s) \times \left[ \min_{f_k} \sum_y P(y|s \cup \{x_n = v_n\}) \cdot MC(f_k, y) \right].
\end{aligned}$$

The test $x_n$ is performed only if its value exceeds its cost, $\text{1-step-VOI}(s, x_n) > C(x_n)$.

Intuitively, the one-step VOI method repeatedly asks the following question: Is it worth executing one more measurement before making a diagnosis, or is it better to make a diagnosis now?

In state $s$, the one-step VOI method first computes the cost of stopping and choosing the action $f_{best}$ that minimizes expected misdiagnosis costs:

$$C(s, f_{best}) = \min_{f_k} \sum_y P(y|s) \cdot MC(f_k, y).$$

Then, for each possible measurement action $x_n$, the method computes the expected cost of measuring $x_n$ and then choosing minimum cost diagnosis actions in each of the resulting

states:

$$\text{1-step-LA}(s, x_n) = C(x_n) + \sum_{v_n} P(x_n = v_n | s) \times \left[ \min_{f_k} \sum_y P(y | s \cup \{x_n = v_n\}) \cdot MC(f_k, y) \right]. \tag{5}$$

Define $x_{best} = \operatorname{argmin}_{x_n} \text{1-step-LA}(s, x_n)$.

With these definitions, we can describe the one-step VOI method in terms of the template in Table 1 as follows. The stopping condition (1) is that $C(s, f_{best}) \leq \text{1-step-LA}(s, x_{best})$; the method also stops when all tests have been performed. The choice of measurement action (2) is $x_{best}$. And the choice of the final diagnosis action in step (3) is $f_{best}$.

## 5. Learning, Overfitting, and Regularization

In the previous section, we considered search algorithms for finding good diagnostic policies. All of these algorithms require various probabilities, particularly $P(x_n = v_n | s)$ and $P(y | s)$ for every state-action pair $(s, x_n)$ or $(s, f_k)$ generated during their search.

One way to obtain these probabilities is to fit a probabilistic model $P(x_1, \ldots, x_N, y)$ to the training data and then apply probabilistic inference to this model to compute the desired probabilities. For example, an algorithm such as K2 (Cooper & Herskovits, 1992) could be applied to learn a Bayesian network from the training data. The advantage of such an approach is that it would cleanly separate the process of learning the probabilities from the process of searching for a good policy.

But the chief disadvantage of such an approach is that it prevents us from exploiting the problem solving task to determine which probabilities should be learned accurately and which probabilities can be ignored (or learned less accurately). Consequently, we have adopted a different approach in which the learning is fully integrated into the search process. This is very important, because it enables us to control overfitting and it also provides additional opportunities for speeding up the search.

The basic way to integrate learning into the search process is very simple. Each time the search algorithm needs to estimate a probability, the algorithm examines the training data and computes the required probability estimate. For example, if an algorithm needs to estimate $P(x_1 = v_1 | \{x_3 = v_3, \ x_5 = v_5\})$, it can make a pass over the training data and count the number of training examples where $x_3 = v_3$ and $x_5 = v_5$. Denote this by $\#(x_3 = v_3, \ x_5 = v_5)$. It can make a second pass over the data and count $\#(x_1 = v_1, \ x_3 = v_3, \ x_5 = v_5)$. From these two quantities, it can compute the maximum likelihood estimate:

$$\hat{P}(x_1 = v_1 \mid \{x_3 = v_3, x_5 = v_5\}) = \frac{\#(x_1 = v_1, \ x_3 = v_3, \ x_5 = v_5)}{\#(x_3 = v_3, \ x_5 = v_5)}.$$

In general,

$$\hat{P}(x_n = v_n | s) = \frac{\#(s \cup \{x_n = v_n\})}{\#(s)}.$$

Similarly, $P(y | s)$ is estimated as the fraction of training examples matching state $s$ that have diagnosis $y$:

$$\hat{P}(y | s) = \frac{\#(s, y)}{\#(s)}.$$

279

This process can obviously be made more efficient by allowing the training data to "flow" through the AND/OR graph (for AO* algorithm) or the classification tree (for greedy algorithms) as it is being constructed. Hence, the starting state (the root) stores a list of all of the training examples. The OR node for state $s$ stores a list of all of the training examples that match $s$. An example *matches* a state if it agrees with all of the measurement results that define that state. An AND node that measures $x_n$ in state $s$ can be viewed as partitioning the training examples stored in OR node $s$ into disjoint subsets according to their observed values on test $x_n$. The same method has been employed in classification tree algorithms for many years (Breiman et al., 1984; Quinlan, 1993).

Unfortunately, this straightforward approach, when combined with both the systematic and greedy search algorithms, often results in overfitting—that is, finding policies that give very good performance on the training data but that give quite poor performance on new cases.

Figure 5 illustrates this for AO*. This figure shows an *anytime graph* in which the value $V^{real}(s_0)$ of the current realistic policy, $\pi^{real}$, is plotted after each node expansion (or iteration of the algorithm). $V^{real}$ is evaluated both on the training data and on a disjoint test data set. On the training data, the quality of the learned policy improves monotonically with the number of iterations—indeed, this is guaranteed by the AO* algorithm. But on the test data, the performance of the realistic policies gets worse after 350 iterations. Upon convergence, AO* has learned the optimal policy with respect to the training data, but this policy performs badly on the test data.

Machine learning research has developed many strategies for reducing overfitting. The remainder of this section describes the regularizers that we have developed for both systematic and greedy search algorithms. First, we discuss regularizers for AO*. Then we discuss regularizers for greedy search.

## 5.1 Regularizers for AO* Search

Overfitting tends to occur when the learning algorithm extracts too much detailed information from the training data. This can happen, for example, when the learning algorithm considers too many alternative policies for a given amount of training data. It can also occur when the algorithm estimates probabilities from very small numbers of training examples. Both of these problems arise in AO*. AO* considers many different policies in a large AND/OR graph. And as the AND/OR graph grows deeper, the probabilities in the deeper nodes are estimated from fewer and fewer training examples.

We have pursued three main strategies for regularization: (a) regularizing the probability estimates computed during the search, (b) reducing the amount of search through pruning or early stopping, and (c) simplifying the learned policy by post-pruning to eliminate parts that may have overfit the training data.

### 5.1.1 LAPLACE CORRECTION (DENOTED BY 'L')

To regularize probability estimates, a standard technique is to employ Laplace corrections. Suppose measurement $x_n$ has $V_n$ possible outcomes. As discussed above, the

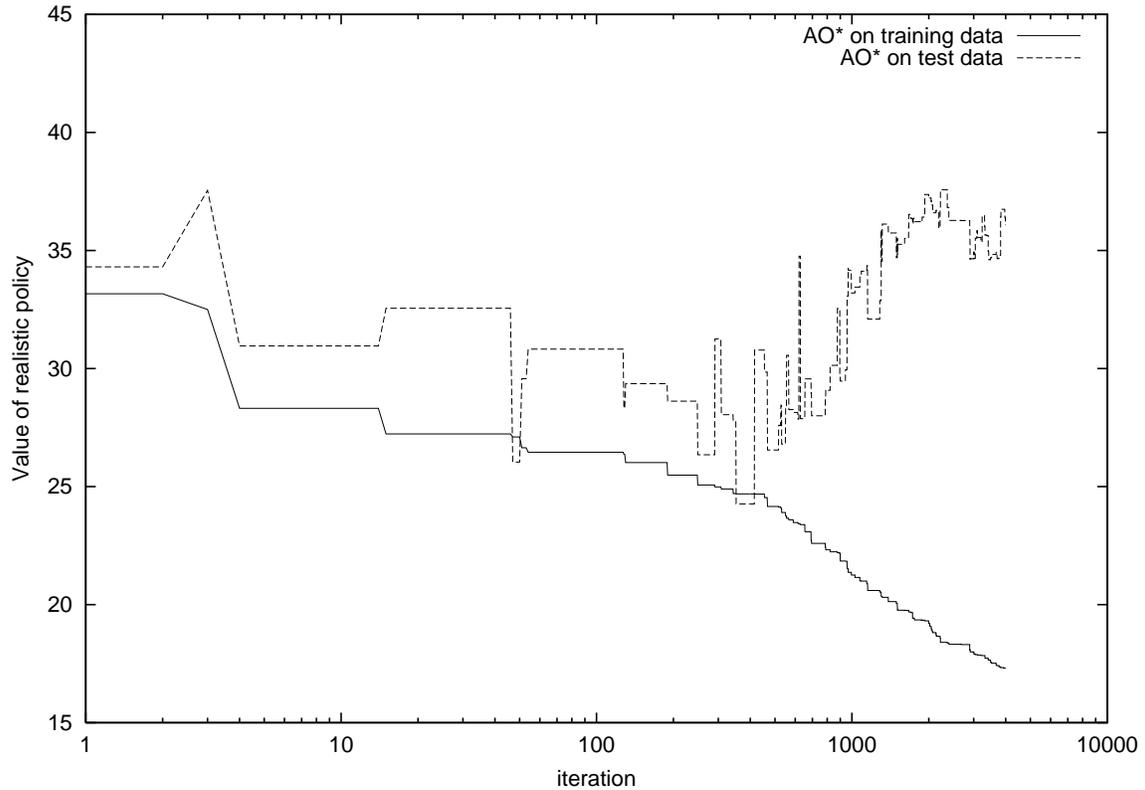

Figure 5: Illustration of AO* overfitting. This anytime graph shows that the best realistic policy, according to the test data, was discovered after 350 iterations, after which AO* overfits.

maximum likelihood estimate for $P(x_n = v_n|s)$ is

$$\hat{P}(x_n = v_n|s) = \frac{\#(s \cup \{x_n = v_n\})}{\#(s)}.$$

The Laplace-corrected estimate is obtained by adding 1 to the numerator and $V_n$ to the denominator:

$$\hat{P}_L(x_n = v_n \mid s) = \frac{\#(s \cup \{x_n = v_n\}) + 1}{\#(s) + V_n}.$$

Similarly, the Laplace-corrected estimate for a diagnosis $y$ is obtained by adding 1 to the numerator and $K$ (the number of possible diagnoses) to the denominator:

$$\hat{P}_L(y|s) = \frac{\#(s, y) + 1}{\#(s) + K}.$$

One advantage of the Laplace correction is that no probability value will ever be estimated as 0 or 1. Those probability values are extreme, and hence, extremely dangerous.

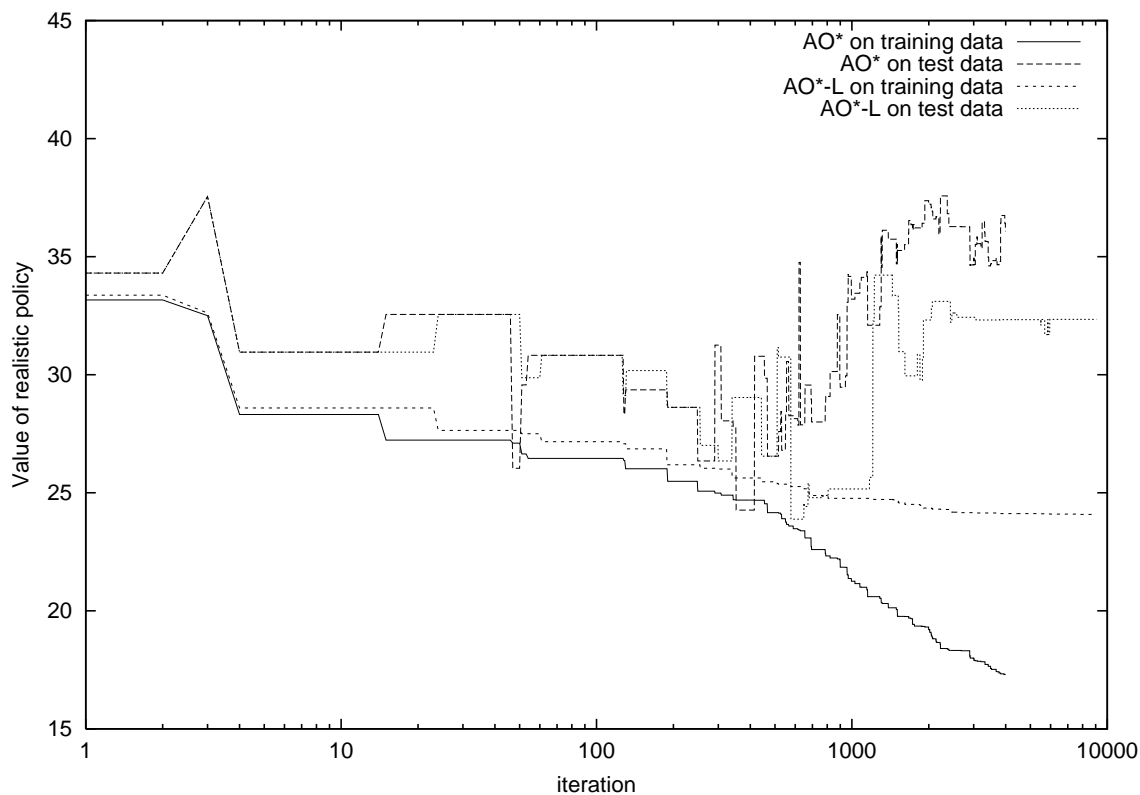

Figure 6: Anytime graphs of AO* and AO* with Laplace correction. The Laplace regularizer helps AO*, both in the anytime graph and in the value of the last policy learned.

For example, if AO* believes that $P(x_n = v_n|s) = 0$, then it will not expand this branch further in the tree. Even more serious, if AO* believes that $P(y = c|s) = 0$, then it will not consider the potential misdiagnosis cost $MC(f_k, y = c)$ when computing the expected costs of diagnosis actions $f_k$ in state $s$.

Figure 6 shows that AO* with the Laplace regularizer gives worse performance on the training data but better performance on the test data than AO*. Despite this improvement, AO* with Laplace still overfits: a better policy that was learned early on is discarded later for a worse one.

### 5.1.2 Statistical Pruning (SP)

Our second regularization technique reduces the size of the AO* search space by pruning subtrees that are unlikely to improve the current realistic policy.

The statistical motivation is the following: given a small training data sample, there are many pairs of diagnostic policies that are statistically indistinguishable. Ideally, we would like to prune all policies in the AND/OR graph that are statistically indistinguishable from the optimal policies. Since this is not possible without first expanding the graph, we need a heuristic that approximately implements the following indifference principle:

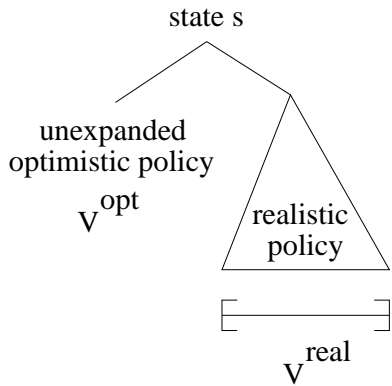

Figure 7: Statistical pruning (SP) checks whether $V^{opt}(s)$ falls inside a confidence interval around $V^{real}(s)$. If it does, then SP prunes $\pi^{opt}(s)$ (the unexpanded optimistic policy).

**Indifference Principle**. *Given two diagnostic policies whose values are statistically indistinguishable based on the training data set, a learning algorithm can choose arbitrarily between them.*

This heuristic is called *statistical pruning* (abbreviated SP), and it is applied in each OR node $s$ whose optimistic policy is selected for expansion. The two diagnostic policies under consideration are the currently unexpanded optimistic policy $\pi^{opt}(s)$ and the current realistic policy $\pi^{real}(s)$. The action specified by $\pi^{opt}(s)$ will be pruned from the graph if a statistical test cannot reject the null hypothesis that $V^{opt}(s) = V^{real}(s)$. In other words, between an incomplete policy $\pi^{opt}$ and a complete policy $\pi^{real}$, we prefer the latter.

The statistical test is computed as follows. To each of the training examples *te* that matches state $s$, we can apply $\pi^{real}$ and compute the total cost of diagnosis (starting from state $s$). From this information, we can compute a 95% confidence interval on $V^{real}(s)$ (e.g., using a standard normal distribution assumption). If $V^{opt}(s)$ falls inside this confidence interval, then we cannot reject the null hypothesis that $V^{opt}(s) = V^{real}(s)$. Therefore, by the indifference principle, we can choose $\pi^{real}(s)$ and prune $\pi^{opt}(s)$. This is illustrated in Figure 7.

Because $V^{opt}(s)$ is a lower bound on $V^*(s)$ (see Theorem 4.1) and $V^{real}(s)$ is an upper bound on $V^*(s)$ (see Theorem 4.2), we can relate statistical pruning to the indifference principle in a slightly stronger way. If $V^{opt}(s)$ falls inside the confidence interval for $V^{real}(s)$, then $V^*(s)$ must also fall inside the confidence interval, because $V^{opt}(s) \leq V^*(s) \leq V^{real}(s)$. Hence, with at least 95% confidence, we cannot reject the null hypothesis that $V^{real}(s) = V^*(s)$. Hence, the indifference principle authorizes us to choose $\pi^{real}$, since it is statistically indistinguishable from the optimal policy. However, this argument only remains true as long as $\pi^{real}$ remains unchanged. Subsequent expansions by AO* may change $\pi^{real}$ and invalidate this statistical decision.

The SP heuristic is applied as the AND/OR graph is grown. When an AND node $(s, a)$ is selected for expansion, SP first checks to see if this AND node should be pruned instead. If it can be pruned, the action $a$ will be ignored in further computations (SP deletes it from the set of available actions $A(s)$). AO* then updates $Q^{opt}, V^{opt}$, and $\pi^{opt}$ in the graph. No updates to $\pi^{real}$ or $V^{real}$ are needed, because pruning $a$ does not change the realistic graph.

In previous work (Bayer-Zubek & Dietterich, 2002), we described a version of the SP heuristic that employed a paired-difference statistical test instead of the simple confidence interval test described here. On synthetic problems, these two statistical tests gave nearly identical results. We prefer the confidence interval test, because it allows us to relate $V^{real}(s)$ and $V^*(s)$.

Some care must be taken when combining statistical pruning with Laplace corrections. With Laplace corrections, the mean of the observed total cost of the training examples matching state $s$ when processed by $\pi^{real}$ is not the same as $V^{real}(s)$, because the latter is computed using Laplace-corrected probabilities. To fix this problem, we compute the width of the confidence interval by applying $\pi^{real}$ to each training example matching state $s$ and then use $V^{real}(s)$ as the center of the confidence interval.

### 5.1.3 EARLY STOPPING (ES)

Another way to limit the size of the search space considered by AO* is to halt the search early. This method has long been applied to regularize neural networks (e.g., Lang, Waibel, & Hinton, 1990). Early stopping employs an internal validation set to decide when to halt AO*. The training data is split in half. One half is called the "subtraining data", and the other half is called the "holdout data". AO* is trained on the subtraining data, and after every iteration, $\pi^{real}$ is evaluated on the holdout data. The $\pi^{real}$ that gives the lowest total cost on the holdout data is remembered, and when AO* eventually terminates, this best realistic policy is returned as the learned policy.

Early stopping can be combined with the Laplace correction simply by running AO* with Laplace corrections on the subtraining set. There is no need to Laplace-correct the evaluation of $\pi^{real}$ on the holdout set.

### 5.1.4 PESSIMISTIC POST-PRUNING (PPP) BASED ON MISDIAGNOSIS COSTS

Our final AO* regularizer is pessimistic post-pruning. It is based on the well-known method invented by Quinlan for pruning classification trees in C4.5 (Quinlan, 1993). PPP takes a complete policy $\pi$ and the training data set and produces a "pruned" policy $\pi'$ with the hope that $\pi'$ exhibits less overfitting. This PPP is applied to the final realistic policy computed by AO*.

The central idea of PPP is to replace the expected total cost $V^\pi(s)$ at each state $s$ with a statistical upper bound $UB(s)$ that takes into account the uncertainty due to the amount and variability of the training data. At internal node $s$, if the upper bound shows that selecting the best diagnosis action would be preferred to selecting measurement action $\pi(s)$, then node $s$ is converted to a leaf node (and the $UB$ estimates of its ancestors in the policy are updated). PPP can be performed in a single traversal of the decision tree for $\pi$.

Computation begins at the leaves of policy $\pi$ (i.e., the states $s$ where $\pi(s)$ chooses a diagnosis action $f_k$). Let $UB(s)$ be the upper limit of a 95% normal confidence interval for $C(s, f_k)$ (i.e., the expected misdiagnosis cost of choosing action $f_k$ in state $s$). This is computed by taking each training example that matches state $s$, assigning it the diagnosis $f_k$, and then computing the misdiagnosis cost $MC(f_k, y)$, where $y$ is the correct diagnosis of the training example.

The upper bound at an internal node is computed according to the recursion

$$UB(s) = C(s, \pi(s)) + \sum_{s'} P_{tr}(s'|s, \pi(s)) \cdot UB(s').$$

This is just the Bellman equation for state $s$ but with the value function replaced by the $UB$ function. $\pi(s)$ will be pruned, and replaced by the diagnosis action $f_{best}$ with the minimum expected cost, if the upper bound on $C(s, f_{best})$ is less than $UB(s)$ for the internal node, computed above. In this case, $UB(s)$ is set to be the upper bound on $C(s, f_{best})$.

PPP can be combined with Laplace regularization as follows. First, in computing $UB(s)$ for a leaf node, $K$ "virtual" training examples are added to state $s$, such that there is one virtual example for each diagnosis. In other words, the normal confidence interval is computed using the misdiagnosis costs of the training examples that match $s$ plus one $MC(\pi(s), y)$ for each possible diagnosis $y$. Note that all probabilities $P(y|s)$ and $P_{tr}(s'|s, \pi(s))$ were already Laplace-corrected when running AO* with Laplace corrections.

### 5.1.5 SUMMARY OF AO* REGULARIZERS

We have described the following regularizers: Laplace corrections (L), statistical pruning (SP), early stopping (ES), and pessimistic post-pruning (PPP). We have also shown how to combine Laplace regularization with each of the others.

## 5.2 Regularizers for Greedy Search

We now describe four regularizers that we employed with greedy search.

### 5.2.1 MINIMUM SUPPORT PRUNING

For the InfoGainCost and InfoGainCost+MC methods, we adopt the minimum support stopping condition of C4.5 (Quinlan, 1993). In order for measurement action $x_n$ to be chosen, at least two of its possible outcomes $v_n$ must lead to states that have at least 2 training examples matching them. If not, then $x_n$ is not eligible for selection in step (2) of Table 1.

### 5.2.2 PESSIMISTIC POST-PRUNING (PPP) BASED ON MISDIAGNOSIS RATES

For the InfoGainCost method, we applied C4.5's standard pessimistic post-pruning. After InfoGainCost has grown the decision tree, the tree is traversed in post-order. For each leaf node $s$, the pessimistic error estimate is computed as

$$UB(s) = n \times \left[ p + z_c \cdot \sqrt{\frac{p(1-p)}{n}} + \frac{1}{2n} \right],$$

where $n$ is the number of training examples reaching the leaf node, $p$ is the error rate committed by the diagnosis action on the training examples at this leaf, and $z_c = 1.15$ is the 75% critical value for the normal distribution. $UB(s)$ is the upper limit of a 75% confidence interval for the binomial distribution $(n, p)$ plus a continuity correction.

At an internal node $s$, the pessimistic error estimate is simply the sum of the pessimistic error estimates of its children. An internal node is converted to a leaf node if the sum of its children's pessimistic errors is greater than or equal to the pessimistic error that it would have if it were converted to a leaf node.

Laplace regularization can be combined with PPP by replacing the observed error rate $p$ with its Laplace-corrected version (this is computed by adding one "virtual" example for each diagnosis).

### 5.2.3 POST-PRUNING BASED ON EXPECTED TOTAL COSTS

For the MC+InfoGainCost method, we apply a post-pruning procedure that is based not on a pessimistic estimate but rather on the estimated total cost of diagnosis. Recall that MC+InfoGainCost grows the decision tree in the same way as InfoGainCost, but it assigns diagnosis actions to the leaf nodes by choosing the action with the smallest expected misdiagnosis cost on the training data.

This can be further regularized by traversing the resulting decision tree and converting an internal node $s$ where $\pi(s) = x_n$ into a leaf node (where $\pi(s) = f_{best}$) if the expected cost of choosing diagnosis action $f_{best}$ is less than the expected total cost of choosing measurement action $x_n$. This is implemented by computing $C(s, f_{best})$ and $Q^\pi(s, x_n)$ and comparing them. If $C(s, f_{best}) \leq Q^\pi(s, x_n)$, then node $s$ is converted to a leaf node. This computation can be carried out in a single post-order traversal of the decision tree corresponding to $\pi$.

Laplace corrections can be combined with this pruning procedure by applying Laplace corrections to all probabilities employed in computing $Q^\pi(s, x_n)$ and $C(s, f_{best})$.

Bradford et al. (1998) present a similar method of pruning decision trees based on misclassification costs (and zero attribute costs), combined with Laplace correction for class probability estimates (there is no Laplace correction for transition probabilities).

It is interesting to note that this post-pruning based on total costs is not necessary for VOI, because pruning is already built-in. Indeed, any internal node $s$ in the VOI policy $\pi$, with $\pi(s) = x_n$, has $Q^\pi(s, x_n) \leq VOI(s, x_n) < C(s, f_{best})$ (the proof of this theorem appears in the thesis of Bayer-Zubek (2003)).

### 5.2.4 LAPLACE CORRECTION

As with AO*, we could apply Laplace corrections to all probabilities computed during greedy search.

For the InfoGainCost method, Laplace correction of diagnosis probabilities $P(y|s)$ does not change the most likely diagnosis. For the MC+InfoGainCost method, Laplace correction of diagnosis probabilities may change the diagnosis action with the minimum expected cost. Laplace correction is not applied in the computation of the information gain $I(x_n; y|s)$. For the InfoGainCost method, Laplace correction is only applied in the pruning phase, to the error rate $p$. For the MC+InfoGainCost method, Laplace correction is applied, as the policy is grown, to $P(y|s)$ when computing $C(s, f_k)$, and it is also applied during the post-pruning based on expected total costs, to both $P(x_n = v_n|s)$ and $P(y|s)$.

For the VOI method, Laplace correction is applied to all probabilities employed in Equation 5 and in the computation of $C(s, f_{best})$.

## 6. Experimental Studies

We now present an experimental study to measure and compare the effectiveness and efficiency of the various search and regularization methods described above. The goal is to identify one or more practical algorithms that learn good diagnostic policies on real problems with modest-sized training data sets. The main questions are: Which algorithm is the best among all the algorithms proposed? If there is no overall winner, which is the most robust algorithm?

### 6.1 Experimental Setup

We performed experiments on five medical diagnosis problems based on real data sets found at the University of California at Irvine (UCI) repository (Blake & Merz, 1998). The five problems are listed here along with a short name in parentheses that we will use to refer to them: Liver disorders (bupa), Pima Indians Diabetes (pima), Cleveland Heart Disease (heart), the original Wisconsin Breast Cancer (breast-cancer), and the SPECT heart database (spect). These data sets describe each patient by a vector of attribute values and a class label. We define a measurement action for each attribute; when executed, the action returns the measured value of that attribute. We define one diagnosis action for each class label.

The domains were chosen for two reasons. First, they are all real medical diagnosis domains. Second, measurement costs have been provided for three of them (bupa, pima, and heart) by Peter Turney (Turney, 1995); for the other two domains, we set all measurement costs to be 1. Table 2 briefly describes the medical domains; more information is available in the thesis of Bayer-Zubek (2003).

Some pre-processing steps were applied to all domains. First, all training examples that contained missing attribute values were removed from the data sets. Second, if a data set contained more than two classes, selected classes were merged so that only two classes (healthy and sick) remained. Third, any existing division of the data into training and test sets was ignored, and the data were simply merged into a single set. Each real-valued attribute $x_n$ was discretized into 3 levels (as defined by two thresholds, $\theta_1$ and $\theta_2$) such that the discretized variable takes on a value of 0 if $x_n \leq \theta_1$, a value of 1 if $\theta_1 < x_n \leq \theta_2$ and a value of 2 otherwise. The values of the thresholds were chosen to maximize the information gain between the discretized variable and the class. The information gain was computed using the entire data set.

For each domain, the transformed data (2 classes, discretized attributes with no missing values) was used to generate 20 random splits into training sets (two thirds of data) and test sets (one third of data), with sampling stratified by class. Such a split (training data, test data) is called a *replica*. We repeated each of our experiments on each of the replicas to obtain a rough idea of the amount of variability that can be expected from one replica to another. However, it is important to note that because the replicas are not independent (i.e., they share data points), we must use caution in combining the results of different replicas when drawing conclusions about the superiority of one algorithm compared to another.

Table 2: Medical domains. For each domain, we list the number of examples, the number of tests, and the minimum and the maximum cost for a test.

| domain | # examples | # tests | min test cost | max test cost |
|---|---|---|---|---|
| bupa | 345 | 5 | 7.27 | 9.86 |
| pima | 768 | 8 | 1 | 22.78 |
| heart | 297 | 13 | 1 | 102.9 |
| breast-cancer | 683 | 9 | 1 | 1 |
| spect | 267 | 22 | 1 | 1 |

### 6.1.1 Setting the Misdiagnosis Costs (MC)

None of the five UCI domains specifies misdiagnosis costs, so we performed our experiments using five different levels of misdiagnosis costs for each domain. These cost levels were designed such that in the initial state $s_0$ both diagnosis decisions $f_0$ and $f_1$ have equal expected cost and so that the diagnostic policies are non-trivial (i.e., they perform at least one measurement, but do not perform all possible measurements). We call the five MC levels MC1, MC2, MC3, MC4, and MC5, and they progressively make misdiagnosis more expensive. Full details of the methodology are given in the thesis of Bayer-Zubek (2003).

### 6.1.2 Memory Limit

For large domains (with many measurements), the AND/OR graph constructed by AO* grows very large, especially in the following cases: the measurements are not very informative; the measurement costs are low relative to the misdiagnosis costs, so our admissible heuristic does not produce many cutoffs; the optimal policy is very deep; and there are many policies tied with the optimal one and AO* needs to expand all of them to prove to itself that there is no better alternative.

To make systematic search feasible, we need to prevent the AND/OR graph from growing too large. We do this by imposing a limit on the total amount of memory that the AND/OR graph can occupy. We measure memory usage based on the amount of memory that would be required by an optimized AND/OR graph data structure. This "theoretical" memory limit is set to 100 MB. Because our actual implementation is not optimized, this translates into a limit of 500 MB. When the memory limit is reached, the current realistic policy is returned as the result of the search. All of our algorithms (greedy and systematic) converge within this memory limit on all five domains, with one exception: AO* with large misdiagnosis costs reaches the memory limit on the spect data set.

It is interesting to note that even on a domain with many measurements, the systematic search algorithms may converge before reaching the memory limit. This is a consequence of the fact that for modest-sized training data sets, the number of reachable states in the MDP (i.e., states that can be reached with non-zero probability by *some* policy) is fairly small, because not all possible combinations of attribute values can appear in a modest-sized data set.

### 6.1.3 Notations for our Learning Algorithms

In the remainder of this paper, we will employ the following abbreviations to identify the various search algorithms and their regularizers. In all cases, the suffix "L" indicates that Laplace corrections were applied to the algorithm as described in Section 5.

- **Nor, Nor-L** denote InfoGainCost with Norton's criterion for selecting actions and pessimistic post-pruning based on misdiagnosis rates.

- **MC-N, MC-N-L** denote MC+InfoGainCost with Norton's criterion for selecting measurement actions. Diagnosis actions are selected to minimize expected misdiagnosis costs. Post-pruning is based on expected total costs.

- **VOI, VOI-L** denote the one-step Value of Information greedy method.

- **AO*, AO*-L** denote AO*.

- **SP, SP-L** denote AO* with Statistical Pruning.

- **ES, ES-L** denote AO* with Early Stopping. For early stopping, half of the training data is held out to choose the stopping point, and the other half is used by AO* to compute transition probabilities.

- **PPP, PPP-L** denote AO* with Pessimistic Post-Pruning.

### 6.1.4 Evaluation Methods

To evaluate each algorithm, we train it on the training set to construct a policy. Then we compute the value of this policy on the test set, which we denote by $V_{test}$. To compute $V_{test}$, we sum the measurement costs and misdiagnosis cost for each test example, as it is processed by the policy, and then divide the total cost for all examples by the number of test examples.

Note that in our framework, $V_{test}$ is always computed using both measurement costs and misdiagnosis costs, even if the policy was constructed by a learning algorithm (e.g., InfoGainCost) that ignores misdiagnosis costs.

In order to compare learning algorithms, we need some way of comparing their $V_{test}$ values to see if there is a statistically significant difference among them. Even if two learning algorithms are equally good, their $V_{test}$ values may be different because of random variation in the choice of training and test data sets. Ideally, we would employ a statistical procedure similar to analysis of variance to determine whether the observed differences in $V_{test}$ can be explained by differences in the learning algorithm (rather than by random variation in the data sets). Unfortunately, no such procedure exists that is suitable for comparing diagnostic policies. Hence, we adopted the following procedure.

As discussed above, we have generated 20 replicas of each of our data sets. In addition, we have built five misdiagnosis cost matrices for each data set. We apply each learning algorithm to each replica using each of the five MC matrices, which requires a total of 500 runs of each learning algorithm for all domains. For each replica and cost matrix and each pair of learning algorithms (call them $alg1$ and $alg2$), we apply the BDELTACOST bootstrap statistical test (Margineantu & Dietterich, 2000) to decide whether the policy constructed by $alg1$ is better than, worse than, or indistinguishable from the policy constructed by $alg2$ (based on a 95% confidence level). Depending on the BDELTACOST results, we say that $alg1$ wins, loses, or ties $alg2$.

The BDELTACOST test is applied to each replica of each data set. These BDELTACOST results are then combined to produce an overall score for each algorithm according to the following *chess metric*. For a given pair of algorithms, $alg1$ and $alg2$, and a domain $D$, let $(wins, ties, losses)$ be the cumulative BDELTACOST results of $alg1$ over $alg2$, across all five misdiagnosis cost matrices and all 20 replicas. The chess metric is computed by counting each win as one point, each tie as half a point, and each loss as zero points:

$$Score(alg1, alg2, D) \stackrel{\text{def}}{=} wins + 0.5 \times ties.$$

We can also compute the *overall chess score* for an algorithm by summing its chess scores against all of the other algorithms:

$$Score(alg1, D) = \sum_{alg2 \neq alg1} Score(alg1, alg2, D).$$

Note that if the total number of "games" played by an algorithm is $Total = wins + ties + losses$, and if all the games turned out to be ties, the chess score would be $0.5 \times Total$, which we will call the Tie-Score. If the algorithm's chess score is greater than the Tie-Score, then the algorithm has more wins than losses.

The pairwise BDELTACOST tests account for variation in $V_{test}$ resulting from the random choice of the test sets. The purpose of the 20 replicas is to account also for random choice of training sets. Ideally, the 20 training sets would be disjoint, and this would allow us to compute an unbiased estimate of the variability in $V_{test}$ due to the training sets. Unfortunately, because the amount of training data is limited, we cannot make the training sets independent, and as a result, the overall chess scores probably underestimate this source of variability.

## 6.2 Results

We now present the results of the experiments.

### 6.2.1 LAPLACE CORRECTION IMPROVES ALL ALGORITHMS

We first studied the effect of the Laplace regularizer on each algorithm. For each of the seven algorithms with the Laplace correction, we computed its chess score with respect to its non-Laplace version, on each domain. The total number of "games" an algorithm plays against its non-Laplace version is 100 (there are 5 misdiagnosis costs and 20 replicas); therefore, Tie-Score = 50.

Figure 8 shows that on each domain, the Laplace-corrected algorithm scores more wins than losses versus the non-Laplace-corrected algorithm (because each score is greater than Tie-Score). This supports the conclusion that the Laplace correction improves the performance of each algorithm. Some algorithms, such as Nor and AO*, are helped more than others by Laplace.

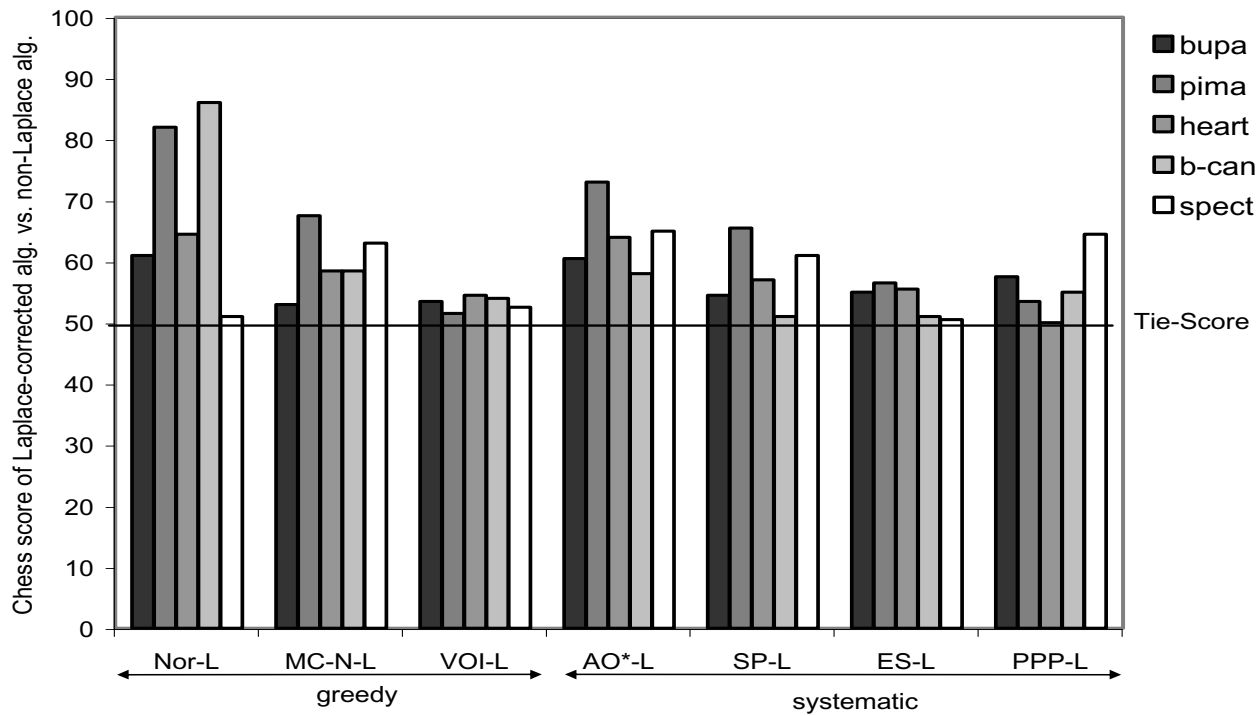

Figure 8: The score of each Laplace-corrected algorithm versus its non-Laplace version, on each domain, is greater than the Tie-Score. Therefore the Laplace version has more wins than losses.

Since the Laplace regularizer improved each algorithm, we decided to compare only the Laplace-corrected versions of the algorithms in all subsequent experiments.

### 6.2.2 The Most Robust Algorithm

To determine which algorithm is the most robust across all five domains, we computed the overall chess score of each Laplace-corrected algorithm against all the other Laplace-corrected algorithms, on each domain. The total number of "games" is 600 (there are 5 misdiagnosis costs matrices, 20 replicas, and 6 "opponent" algorithms); therefore, the Tie-Score is 300.

Figure 9 shows that the best algorithm varies depending on the domain: ES-L is best on bupa, VOI-L is best on pima and spect, SP-L is best on heart, and MC-N-L is best on breast-cancer. Therefore no single algorithm is best everywhere. Nor-L is consistently bad on each domain; its score is always below the Tie-Score. This is to be expected, since Nor-L does not use misdiagnosis costs when learning its policy. MC-N-L, which does use misdiagnosis costs, always scores better than Nor-L.

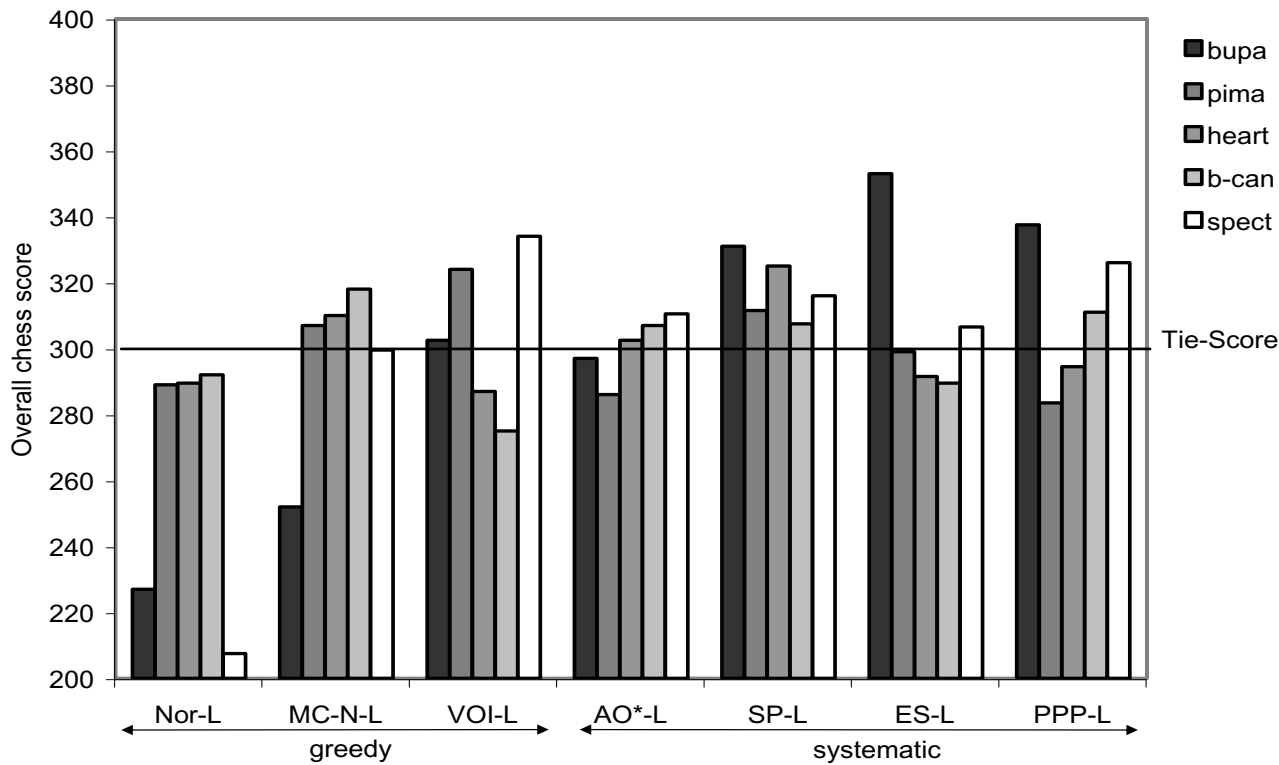

Figure 9: The overall chess score of each Laplace-corrected algorithm, versus all the other Laplace-corrected algorithms. The most robust algorithm is SP-L; it was the only one whose score is greater than Tie-Score (and therefore it has more wins than losses) on every domain.

The fact that VOI-L is best in two domains is very interesting, because it is an efficient greedy algorithm. Unfortunately, VOI-L obtains the worst score in two other domains: heart and breast-cancer.

The only algorithm that has more wins than losses in every domain is SP-L, which combines AO* search, Laplace corrections, and statistical pruning. SP-L always scored among the top three algorithms. Consequently, we recommend it as the most robust algorithm. However, in applications where SP-L (or any of the systematic search algorithms) is too expensive to run, VOI-L can be recommended, since it is the best of the greedy methods.

In addition to looking at the overall chess scores, we also studied the actual $V_{test}$ values. To visualize the differences in $V_{test}$ values, we plotted a graph that we call a "pair graph". Figure 10 shows pair graphs comparing VOI-L and SP-L on all five domains. The horizontal axis in each graph corresponds to the 20 replicas, and the vertical axis to $V_{test}$ values for the two algorithms (VOI-L and SP-L) on that replica. The 20 replicas are sorted according to the $V_{test}$ of VOI-L. If the two algorithms were tied on a replica (according to BDELTACOST), then their $V_{test}$ values are connected by a vertical dotted line.

On bupa and heart, the $V_{test}$ of SP-L is mostly smaller (better) than the $V_{test}$ of VOI-L, but BDELTACOST finds them tied. On pima and spect, the situation is reversed (VOI-L is almost always better than SP-L), and on several replicas the difference is statistically significant. On breast-cancer, SP-L is better than VOI-L, and again the difference is sometimes significant. In general, the pair graphs confirm the chess score results and support our main conclusion that SP-L is the most robust learning algorithm.

### 6.2.3 IMPACT OF HEURISTICS AND REGULARIZERS ON MEMORY CONSUMPTION

We now consider the effect of the admissible heuristic and the Laplace and Statistical Pruning regularizers on the amount of memory required for AO* search. To do this, we measured the amount of memory consumed by five different algorithm configurations: AO* without the admissible heuristic, AO* with the admissible heuristic, AO* with the admissible heuristic and the Laplace correction, AO* with the admissible heuristic and statistical pruning, and, finally, AO* with the admissible heuristic, Laplace correction, and statistical pruning. For AO* without the admissible heuristic, we set the action-value of every unexpanded AND node $(s, x_n)$ to zero, i.e., $Q^{opt}(s, x_n) = 0$. The results are plotted in Figure 11. The memory amounts plotted are computed by taking the actual memory consumed by our implementation and converting it to the memory that would be consumed by an optimized implementation.

There are several important conclusions to draw from these figures. First, note that AO* without the admissible heuristic requires much more memory than AO* with the admissible heuristic. Hence, the admissible heuristic is pruning large parts of the search space. This is particularly evident at low settings of the misdiagnosis costs (MC1 and MC2). At these low settings, AO* is able to find many cutoffs because the expected cost of diagnosis is less than the cost of making additional measurements (as estimated by the admissible heuristic). The savings is much smaller at MC levels 4 and 5.

The second important conclusion is that the Laplace correction increases the size of the search space and the amount of memory consumed. The reason is that without the Laplace correction, many test outcomes have zero probability, so they are pruned by AO*. With the Laplace correction, these outcomes must be expanded and evaluated. The effect is very minor at low MC levels, because the AND/OR graph is much smaller, and consequently there is enough training data to prevent zero-probability outcomes. But at high MC levels, the Laplace correction can cause increases of a factor of 10 or more in the amount of memory consumed.

The third important conclusion is that statistical pruning significantly decreases the size of the AND/OR graph in almost all cases. The only exception is heart at MC4 and MC5, where statistical pruning increases the amount of memory needed by AO*. It seems paradoxical that statistical pruning could lead to an overall increase in the size of the AND/OR graph that is explored. The explanation is that there can be an interaction between statistical pruning at one point in the AND/OR graph and additional search elsewhere. If a branch is pruned that would have given a significantly smaller value for $V^{\pi}$, this can pre-

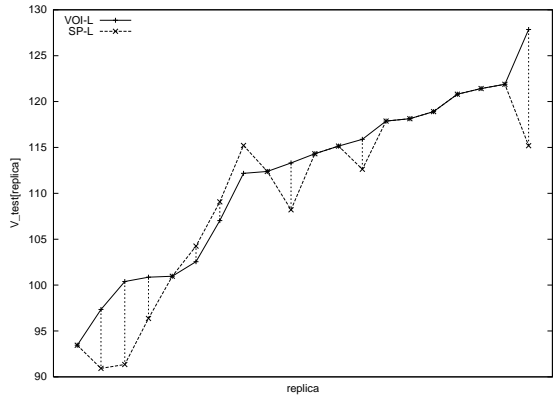

(a) bupa

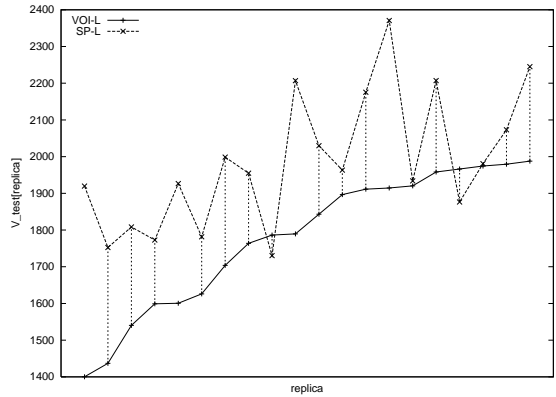

(b) pima

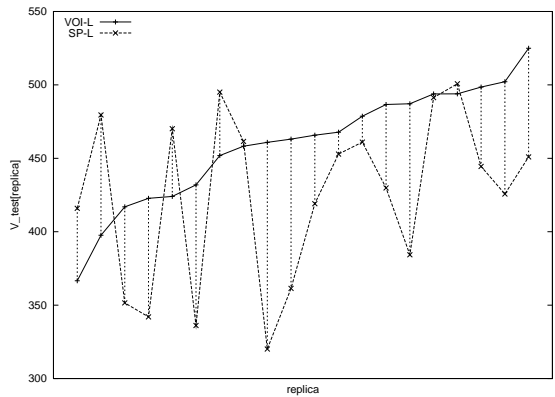

(c) heart

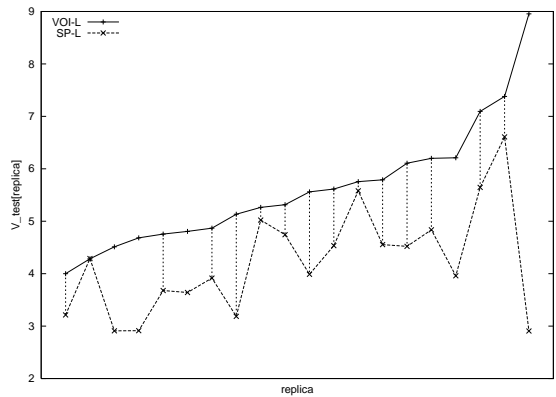

(d) breast-cancer

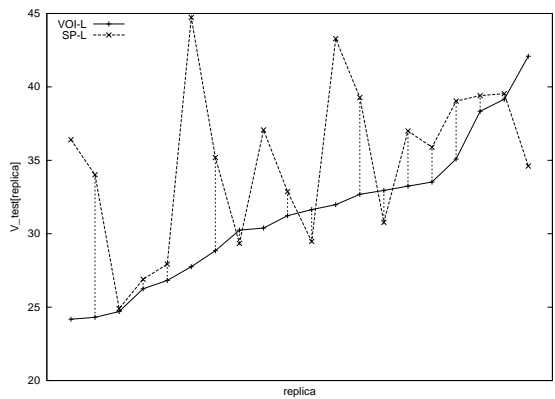

(e) spect

Figure 10: Pair graphs for VOI-L and SP-L for every domain and replica, for the largest misdiagnosis costs MC5. The replicas are sorted by increasing order of $V_{test}$ for VOI-L. Vertical lines connect $V_{test}$ values that are tied according to BDELTA-COST.

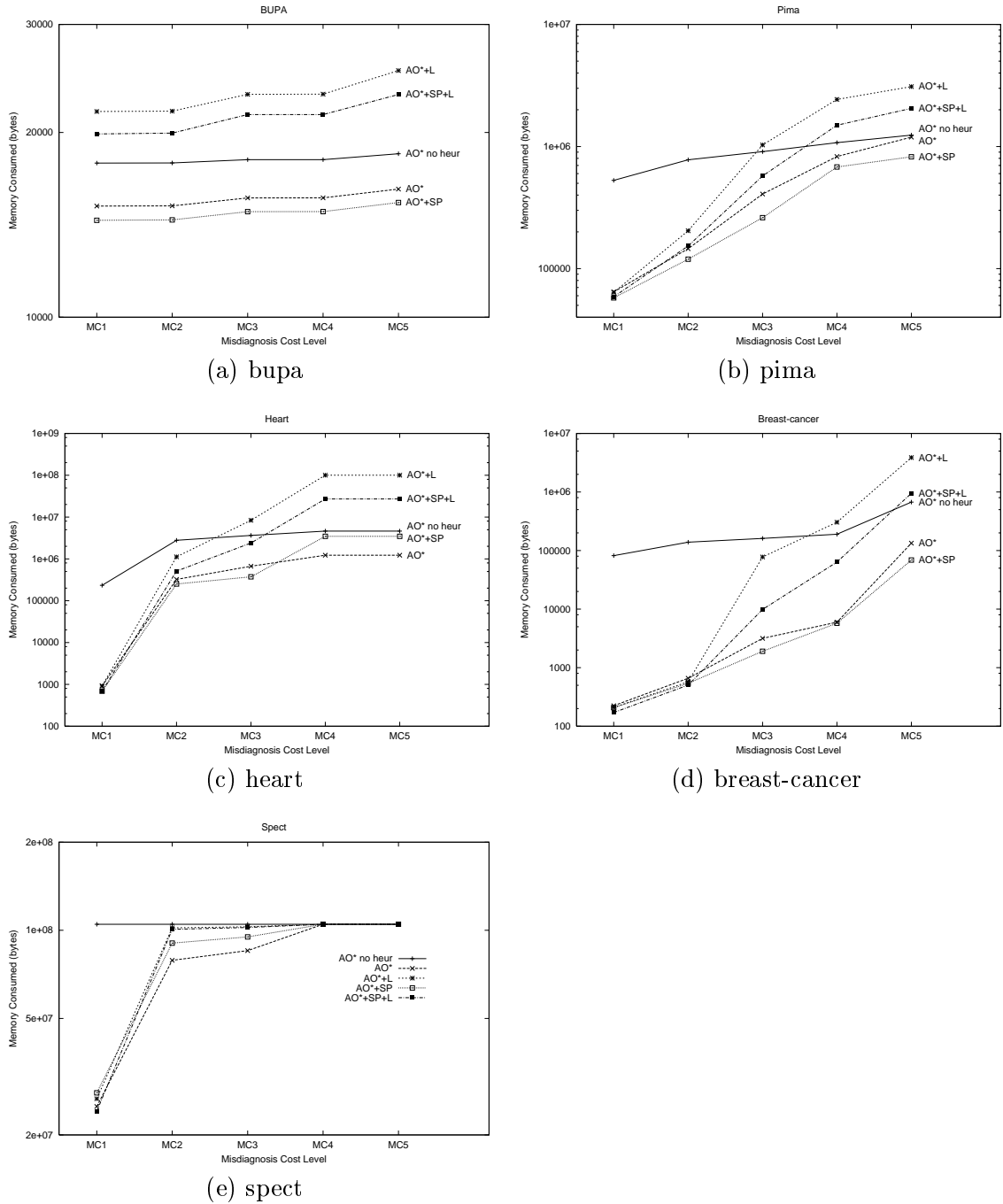

Figure 11: Memory consumed in each domain for five combinations of AO* with and without the admissible heuristic, Laplace corrections, and statistical pruning and for five levels of misdiagnosis costs.

vent heuristic cutoffs elsewhere in the graph. So in some cases, pruning can increase overall memory consumption.

The final conclusion is that statistical pruning dramatically reduces the amount of memory required by AO* with the Laplace correction. Even in cases, such as heart, where statistical pruning causes AO* (without Laplace) to consume more space, SP reduces the amount of space needed with the Laplace correction by nearly an order of magnitude. Nonetheless, statistical pruning is not able to completely compensate for the extra memory consumption of the Laplace corrections, so the final algorithm (AO* + SP + L) requires more memory than AO* without any admissible heuristic at high MC levels, and AO* + SP + L requires much more memory than AO* with the admissible heuristic.

Despite the large amount of memory required, there was only one domain (spect at MC4 and MC5) where the AO* hit the memory limit. Hence, we can see that in terms of memory, systematic search with AO* is feasible on today's desktop workstations.

### 6.2.4 CPU TIME

In addition to measuring memory consumption, we also measured the CPU time required by all of our algorithms. The results are plotted in Figure 12. As expected, the systematic search algorithms require several orders of magnitude more CPU time than the greedy methods. However, even the most expensive algorithm configurations require less than 1000 seconds to execute. Note that as the misdiagnosis cost level increases, the amount of CPU time increases. This is a direct reflection of the corresponding increase in the size of the AND/OR graph that is explored by the algorithms.

## 7. Conclusions

The problem addressed in this paper is to learn a diagnostic policy from a data set of labeled examples, given both measurement costs and misdiagnosis costs. The tradeoff between these two types of costs is an important issue that machine learning research has only just begun to study.

We formulated the process of diagnosis as a Markov Decision Problem. We then showed how to apply the AO* algorithm to solve this MDP to find an optimal diagnostic policy. We also showed how to convert the AO* algorithm into an anytime algorithm by computing the realistic policy at each point in the search (the realistic policy is the best complete policy found so far). We defined an admissible heuristic for AO* that is able to prune large parts of the search space on our problems. We also presented three greedy algorithms for finding diagnostic policies.

The paper then discussed the interaction between learning from training data and searching for a good diagnostic policy. Experiments demonstrated that overfitting is a very serious problem for AO*. The central contribution of the paper is the development of methods for regularizing the AO* search to reduce overfitting and, in some cases, also to reduce the size of the search space. Four regularization techniques (Laplace corrections, statistical pruning, early stopping, and pessimistic post-pruning) were presented. The paper also introduced regularizers for the greedy search algorithms by extending existing methods from classification tree learning.

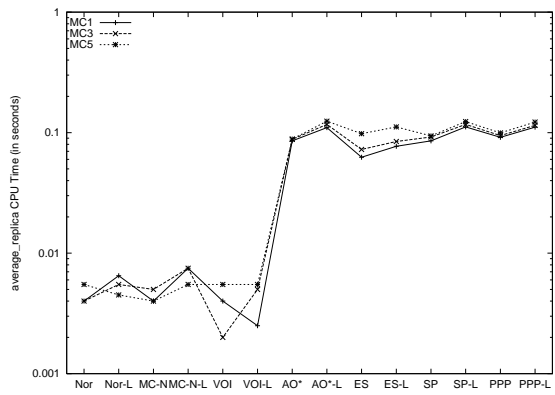

(a) bupa

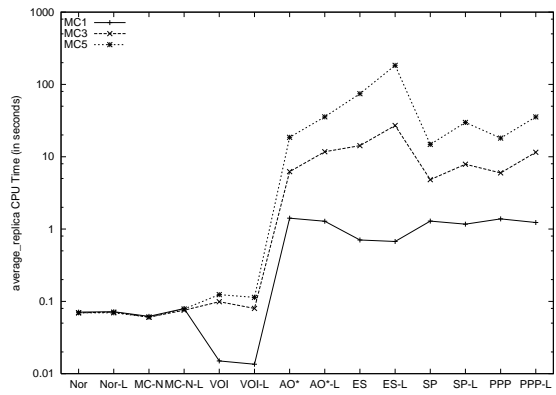

(b) pima

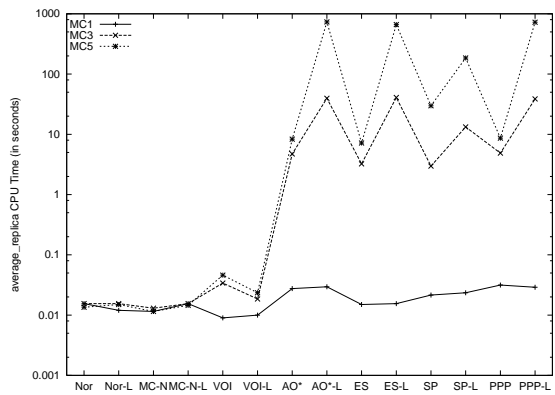

(c) heart

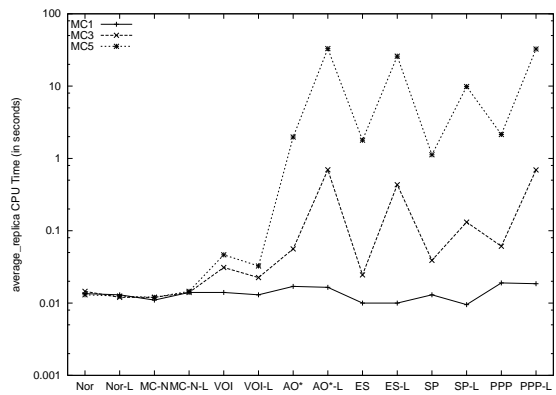

(d) breast-cancer

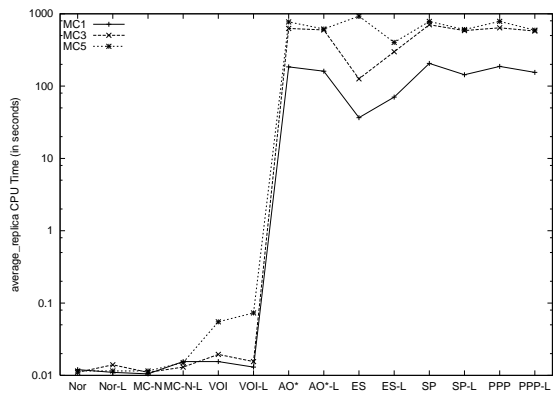

(e) spect

Figure 12: CPU time for all 14 algorithm configurations on the five domains (in each case averaged over 20 replicas). The three curves plot CPU time for misdiagnosis cost levels MC1, MC3, and MC5.

The various search and regularization algorithms were tested experimentally on five classification problems drawn from the UCI repository. A methodology for assigning misdiagnosis costs was developed so that these problems could be converted into cost-sensitive diagnosis problems. The paper also introduced a methodology for combining the results of multiple training/test replicas into an overall "chess score" for evaluating the learning algorithms.

The experiments showed that all of the search algorithms were improved by including Laplace corrections when estimating probabilities from the training data. The experiments also showed that the systematic search algorithms were generally more robust than the greedy search algorithms across the five domains. The best greedy algorithm was VOI-L, but although it obtained the best score on two domains, it produced the worst score on two other domains. The most robust learning algorithm was SP-L. It combines systematic AO* search with Laplace corrections and statistical pruning.

Systematic search for diagnostic policies has not been studied previously by machine learning researchers, probably because it has generally been regarded as computationally infeasible. A surprising conclusion of this paper is that AO* is computationally feasible when applied to the problem of learning diagnostic policies from training examples. This conclusion is based both on experimental evidence—AO* required less than 500 MB of memory on virtually all of our benchmark scenarios—and theoretical analysis.

From a theoretical perspective, there are five factors that help make AO* feasible in this setting: the modest amount of training data, the modest number of possible tests, the small number of outcomes for each test, the admissible heuristic, and the statistical pruning regularizer. We discuss each of these factors in turn:

**Modest amount of training data.** In learning for diagnosis, there is a cost for measuring each attribute of each training example. Consequently, each training example is expensive to collect, and this puts a practical limit on the size of the training data set. This in turn limits the space of reachable states in the MDP. As a result, the AND/OR graph searched by AO* does not grow too large. As the amount of training data grows, this graph will gradually grow larger and at some point, it will become too large for the available memory. Good results may still be obtained by imposing a memory limit, as we did in the spect experiments.

**Modest number of possible tests.** Our experiments only considered domains with 22 or fewer tests. The number of tests determines the branching factor of the OR nodes in the graph, so the size of the graph scales exponentially in this quantity. However, if most of the tests can be pruned by the admissible heuristic or by statistical pruning, the exponential explosion can be avoided. Whether this is possible in any particular problem depends on the relative costs and informativeness of the different tests.

**Small number of outcomes for each test.** We discretized each continuous measurement to have only 3 outcomes. The number of outcomes determines the branching factor of the AND nodes in the graph, so the graph size scales exponentially in this quantity as well. This quantity can be controlled through discretization (see below).

**The admissible heuristic.** The problem of learning for diagnosis is non-trivial only when the costs of making measurements are comparable to the costs of misdiagnosis. But when this is true, our admissible heuristic is able to prune large parts of the search space.

**Statistical pruning.** Finally, the statistical pruning regularizer is able to prune parts of the search space that are unlikely to produce improved policies.

Notice that the size of the AND/OR graph does not increase as the number of possible diagnoses increases. Hence, the AO* search approach scales well with the number of possible diagnostic outcomes.

In cases where the AND/OR graph becomes infeasibly large, we recommend VOI-L, since our experiments showed that it was the best greedy method.

The MDP framework for diagnosis is general enough to handle several extensions to the learning algorithms studied in this paper. For example, in our experiments, we considered only diagnosis problems that involve two classes, "healthy" and "sick." But this could easily be generalized to consider an arbitrary number of classes. Our implementations assumed that the cost of a measurement depends only on the attribute being measured, $C(x_n)$. This can easily be generalized so that the cost of a measurement depends on which tests have already been executed and the results that they produced, and it can also depend on the result of the measurement. In other words, the cost function of a measurement can be generalized to $C(s, x_n, s')$, where $s$ is the current state of the MDP, $x_n$ is the measurement, and $s'$ is the resulting state $s' = s \cup \{x_n = v_n\}$. Our implementations also assumed that the misdiagnosis costs were fixed for all patients, but this could be extended to allow the costs to vary from one patient to another. These changes in the diagnosis problem (multiple classes and complex costs) do not modify the size or complexity of the MDP.

Some important extensions to the diagnostic setting will require extensions to the MDP framework as well. For example, to handle treatment actions that have side effects, noisy actions that may need to be repeated, or actions that have delayed results, the definition of a state in the MDP needs to be extended. An initial examination of these extensions suggest that each of them will cause the MDP state space to grow significantly, and this may make it infeasible to search the space of diagnostic policies systematically. Hence, these extensions will probably require new ideas for their solution.

Another important direction for future work is to extend our approach to handle tests with a large number of possible outcomes, including particularly tests with continuous measured values. We applied standard information-gain methods for discretizing continuous attributes, but an interesting direction for future work is to develop cost-sensitive discretization methods.

A final challenge for future research is to learn good diagnostic policies from incomplete training data. The algorithms presented in this paper assume that each attribute of each training example has been measured. Such data are hard to obtain. But every day, thousands of patients are seen by physicians, medical tests are performed, and diagnostic decisions are made. These data are incomplete, because each physician is following his or her own diagnostic policy that certainly does not perform all possible medical tests. The resulting training examples have many missing values, but these values are not "missing at random", so standard methods for handling missing values cannot be applied. Methods for learning diagnostic policies from such data would be very valuable in many applications.

The problem of learning a diagnostic policy from data collected while executing some other diagnostic policy is identical to the problem of "off-policy" reinforcement learning (Sutton & Barto, 1999). In reinforcement learning, the diagnostic policy generating the data is called the exploration policy. Much is known about creating exploration policies that enable learning of optimal policies. For example, if the exploration policy has non-zero probability of executing every action in every state, then the optimal policy can still be learned. If the exploration policy can be controlled by the learning system, then much more selective exploration can produce the optimal policy (Kearns & Singh, 1998). By extending these ideas, it may be possible to learn diagnostic policies from data collected routinely in hospitals and clinics.

The problem of learning diagnostic policies is fundamental to many application domains including medicine, equipment diagnosis, and autonomic computing. A diagnostic policy must balance the cost of gathering information by performing measurements with the cost of making incorrect diagnoses. This paper has shown that AO*-based systematic search, when combined with regularization methods for preventing overfitting, is a feasible method for learning good diagnostic policies from labeled examples.

## Acknowledgments

The authors gratefully acknowledge the support of the National Science Foundation under grants IRI-9626584, EIA-9818414, IIS-0083292, EIA-0224012, and ITR-5710001197. The authors also gratefully acknowledge the support of the Air Force Office of Scientific Research under grant number F49620-98-1-0375.

This paper extends the conference paper of Bayer-Zubek (2004).

The authors thank the anonymous reviewers for their comments.